\documentclass[10pt,twocolumn,letterpaper]{article}

\usepackage[dvipsnames,table,xcdraw]{xcolor}
\usepackage{cvpr}              % To produce the CAMERA-READY version

\usepackage[title]{appendix}

\usepackage{graphicx}
\usepackage{amsmath}
\usepackage{bm}
\usepackage{amssymb}
\usepackage{amsbsy}
\usepackage{fixmath}
\usepackage{booktabs}
\usepackage{setspace}
\usepackage{pifont}% http://ctan.org/pkg/pifont
\newcommand{\cmark}{\textcolor{green}{\ding{51}}}%
\newcommand{\xmark}{\textcolor{red}{\ding{55}}}%

\usepackage[normalem]{ulem}
\newcommand{\stkout}[1]{\ifmmode\text{\uwave{\ensuremath{#1}}}\else\uwave{#1}\fi}

\usepackage[numbers,sort&compress]{natbib}

\usepackage{float}
\usepackage{xspace}
\usepackage{epsfig}
\usepackage[resetlabels,labeled]{multibib}

\usepackage{blindtext}
\usepackage{amsmath,amsfonts,bm}

\def\eqref#1{equation~\ref{#1}}
\def\1{\bm{1}}

\def\rvg{{\mathbf{g}}}

\def\rvx{{\mathbf{x}}}
\def\rvy{{\mathbf{y}}}
\def\rvz{{\mathbf{z}}}

\def\ervg{{\textnormal{g}}}

\def\rmW{{\mathbf{W}}}

\DeclareMathAlphabet{\mathsfit}{\encodingdefault}{\sfdefault}{m}{sl}
\SetMathAlphabet{\mathsfit}{bold}{\encodingdefault}{\sfdefault}{bx}{n}

\usepackage{nccmath}

\usepackage{multirow,booktabs}
\usepackage{colortbl}
\usepackage{tabularx}
\usepackage{url}
\usepackage{algorithm,algpseudocode}

\usepackage{enumitem}
\setlist{topsep=0pt, leftmargin=*}
\usepackage[pagebackref,breaklinks,colorlinks]{hyperref}

\usepackage{fontawesome5}

\usepackage[capitalize]{cleveref}
\crefname{section}{Sec.}{Secs.}
\Crefname{section}{Section}{Sections}
\Crefname{table}{Table}{Tables}
\crefname{table}{Tab.}{Tabs.}

\definecolor{myblue}{RGB}{150, 168, 207}
\definecolor{mypink}{rgb}{0.89, 0.0, 0.13}
\definecolor{myred}{RGB}{255,33,33}
\definecolor{mygreen}{rgb}{0.0, 0.6, 0.0}

\definecolor{fusionblue}{RGB}{140,167,206}
\definecolor{fusionpink}{RGB}{202,122,118}
\definecolor{mypurple}{RGB}{150,115,166}

\def\cvprPaperID{185} % *** Enter the CVPR Paper ID here
\def\confName{CVPR}
\def\confYear{2022}

\newcommand{\method}{SplitNets\xspace}

\newcommand{\mysp}{splitting point\xspace}

\newcommand{\fakeparagraph}[1]{\vspace{1mm}\noindent\textbf{#1}}

\newcommand{\tabcheck}{~~~\llap{\makebox[0pt][l]{$\square$}\raisebox{.15ex}{\hspace{0.1em}$\checkmark$}}~}
  {}% \end{boldenv}

\begin{document}

\title{\method: Designing Neural Architectures \\ for Efficient Distributed Computing on Head-Mounted Systems}

\author{Xin Dong$^{1}$, 
Barbara De Salvo$^{2}$, 
Meng Li$^{2}$,
Chiao Liu$^{2}$, 
Zhongnan Qu$^{3}$,
H.T. Kung$^{1}$ 
and Ziyun Li$^{2}$\\
\vspace{-0.35cm}
\\
% $^1$Princeton University, $^2$UIUC, $^3$NVIDIA\\
$^1$Harvard University,\quad$^2$Meta Reality Labs,\quad$^3$ETH Zurich\\
\tt\small xindong@g.harvard.edu
}
\maketitle

%%%%%%%%% ABSTRACT
\vspace{-1em}
\begin{abstract}
We design deep neural networks (DNNs) and corresponding networks' splittings to distribute DNNs' workload to camera sensors and a centralized aggregator on head mounted devices to meet system performance targets in inference accuracy and latency under the given hardware resource constraints. To achieve an optimal balance among computation, communication, and performance, a split-aware neural architecture search framework, \method, is introduced to conduct model designing, splitting, and communication reduction simultaneously. We further extend the framework to multi-view systems for learning to fuse inputs from multiple camera sensors with optimal performance and systemic efficiency. We validate \method for single-view system on ImageNet as well as multi-view system on 3D classification, and show that the \method framework achieves state-of-the-art (SOTA) performance and system latency compared with existing approaches.

\end{abstract}

%%%%%%%%% BODY TEXT
\vspace{-1em}
\section{Introduction}
\label{sec:intro}

\begin{figure}[t]
    \centering
    \includegraphics[width=0.9\linewidth]{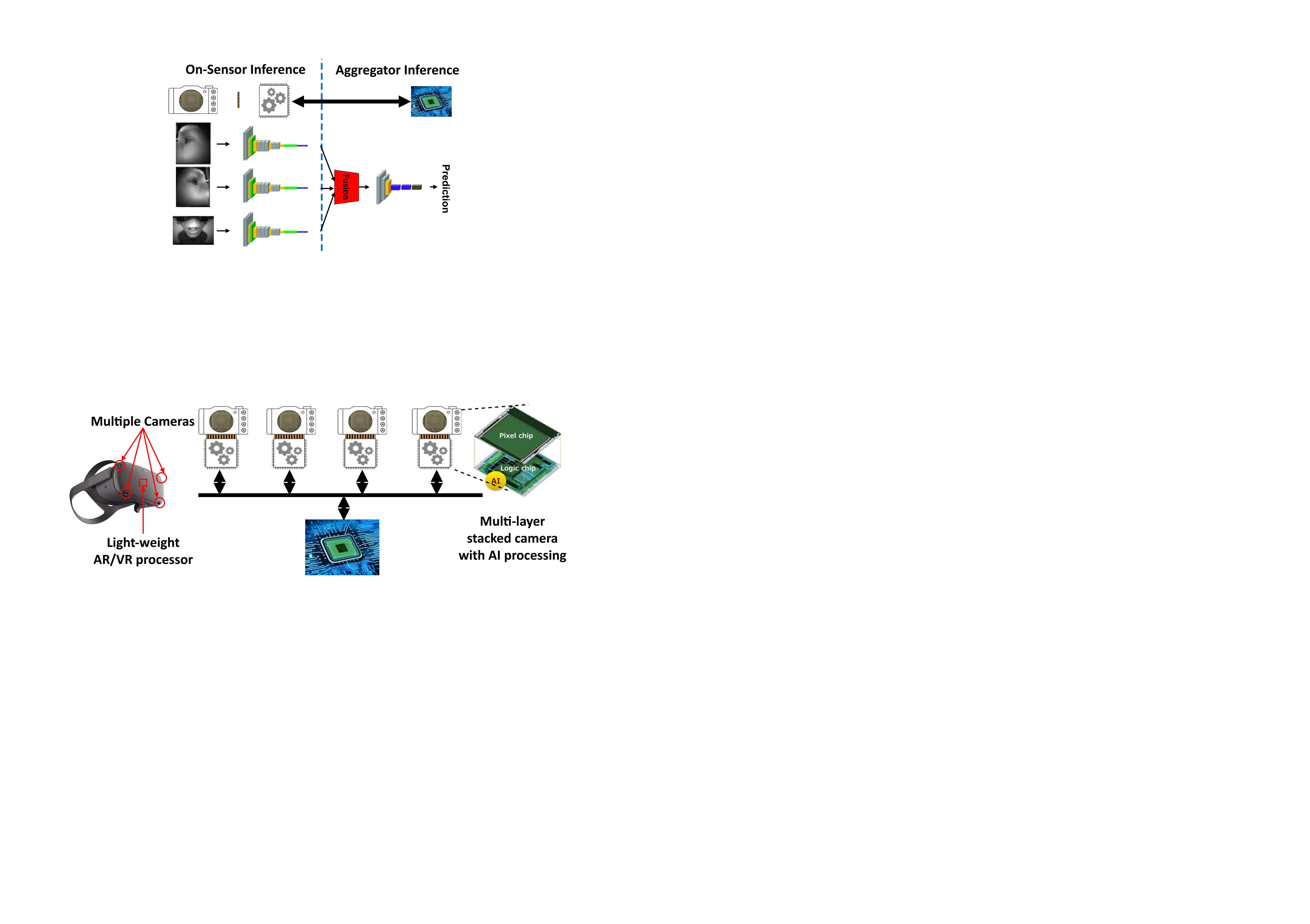}
    \caption{AR/VR device with multiple intelligent AI cameras.}
    \label{fig:system_archi}
    \vspace{-0.8em}
\end{figure}

\begin{figure}[t]
    \centering
    \includegraphics[width=0.75\linewidth]{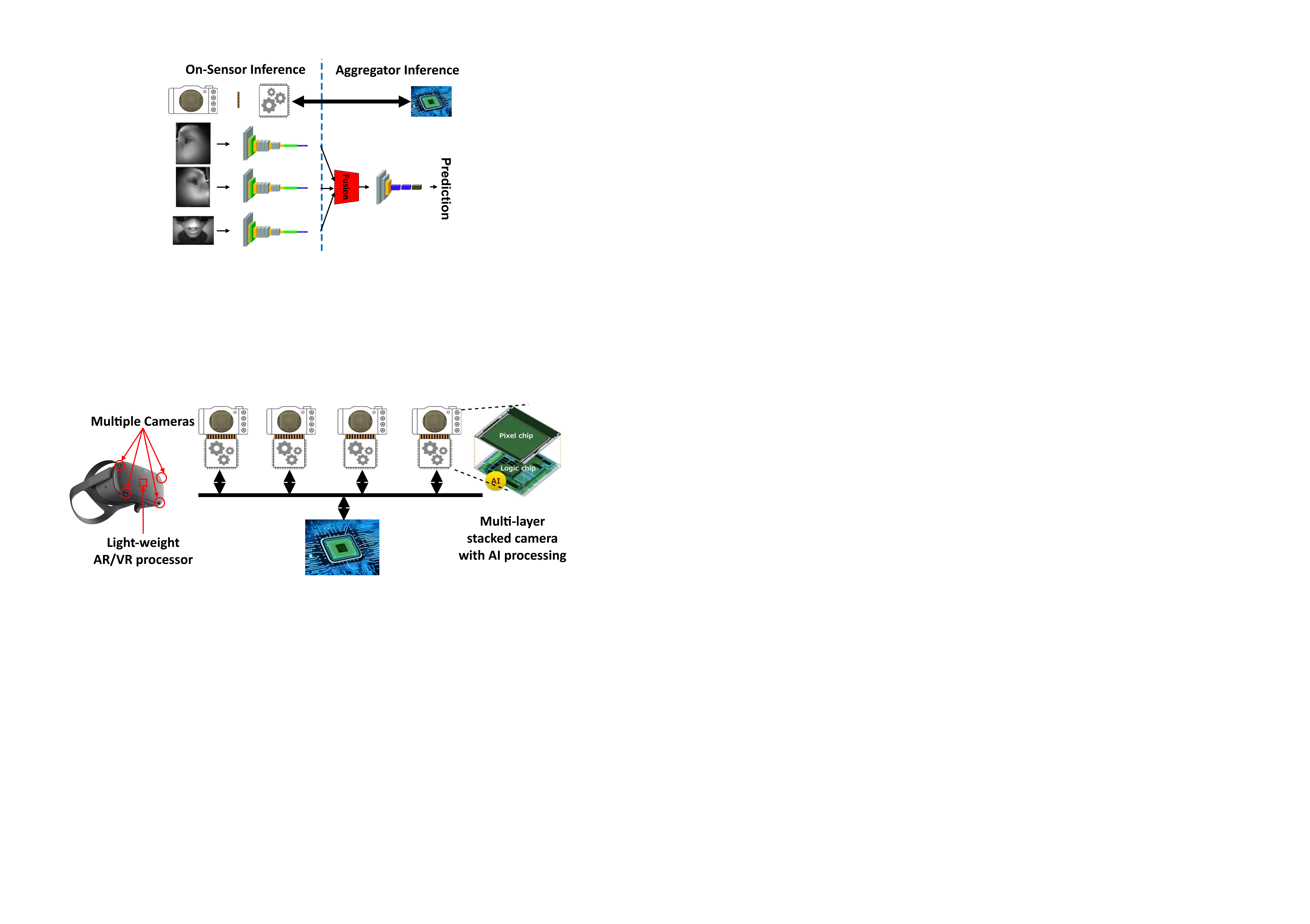}
    \caption{{\small Distributed inference with sensors and an aggregator. In this work, we focus on single/multi-view classification tasks.}}
    \label{fig:compute_partition}
    \vspace{-0.8em}
\end{figure}

\begin{table*}[!ht]
    \small
    \centering
    \begin{tabular}{c|ccccc}
    \toprule
       Advantages & $\boldsymbol{\downarrow}$ Comm. Cost & $\boldsymbol{\downarrow}$ Peak Sensor Mem.  & $\boldsymbol{\uparrow}$ Parallelism & $\boldsymbol{\uparrow}$ Hardware Utilization & Privacy Preserving \\
       \midrule
       All on sensors  &  \cmark & \xmark & \cmark & \xmark & \cmark \\
       All on aggregator &  \xmark & \cmark & \xmark & \xmark & \xmark\\
       Distributed computing & \cmark & \cmark & \cmark & \cmark & \cmark\\
    \bottomrule
    \end{tabular}
    \vspace{-1mm}
    \caption{Comparison of different DNN computing offload paradigms.}
    \label{tab:compare_paradigm}
    \vspace{-3mm}
\end{table*}

\begin{table}[ht]
    % \centering
    \small
    \resizebox{\linewidth}{!}{
    \begin{tabular}{l|l}
    \toprule
    ~~Components & \multicolumn{1}{c}{System factors to be considered} \\
    \midrule
    \faEye[regular]\quad Sensor & \begin{tabular}[l]{@{}l@{}} 
    \tabcheck Computation capability \\
    % \tabcheck Energy consumption \\
    \tabcheck {Peak memory constraint} \\
    \tabcheck Number of sensors (\ie, parallelism)\\
    % \tabcheck \underline{Each sensor only processes its own images}\\
    \end{tabular}\\
    \midrule
    \faNetworkWired\quad Comm. & \begin{tabular}[l]{@{}l@{}} 
    % \tabcheck Energy consumption \\
    % \tabcheck Communication speed \\
    \tabcheck Communication bandwidth \\
    \end{tabular}\\
    \midrule
    \faMicrochip~Aggregator & \begin{tabular}[l]{@{}l@{}} 
    \tabcheck Computation capability \\
    % \tabcheck Energy consumption \\
    \end{tabular}\\
    \midrule
    \midrule
    Whole system & \begin{tabular}[l]{@{}l@{}} 
    \tabcheck Task performance (\eg, accuracy) \\
    \tabcheck Overall latency \textbf{=} on-sen. latency \\~\quad\quad\quad~~~~\textbf{+} comm. latency \textbf{+} on-agg. latency\\
    \end{tabular}\\
    \bottomrule
    \end{tabular}
    }
    \vspace{-1mm}
    \caption{Factors need to be balanced when distributing DNN computation. The listed factors are interwoven with each other and have to be considered holistically. 
    % Underlined factors are hard constraints.
    }
    \label{tab:system_factors}
    \vspace{-3mm}
\end{table}

Virtual Reality (VR) and Augmented Reality (AR) are becoming increasingly prevailing as one of the next-generation computing platforms~\cite{michealiedm2021}. Head mounted devices (HMD) for AR/VR feature multiple cameras to support various computer vision (CV) / machine learning (ML) powered human-computer interaction functions, such as object classification~\cite{qin2021virtual,zheng2020four}, hand-tracking~\cite{han2020megatrack} and SLAM~\cite{mur2015orb}. 

Due to the recent advances of camera technologies, tiny multi-layer stacked cameras with AI computing capability arise~\cite{liu2020intelligent,liu2019intelligent}, as depicted in~\Cref{fig:system_archi}. Because of the small form factor, these intelligent cameras have highly constrained resources compared with general purpose mobile or data center grade AI processors. However, each camera is still capable of performing pre-processing directly after image acquisition, significantly reducing expensive raw image data movement. In a modern HMD system, the distributed intelligent stacked image sensors and a central AR/VR processor (aggregator) form the hardware backbone to realize complex CV/ML functions on device~\cite{michealiedm2021,pinkham2021near,rub_takl}.
% However, each of these cameras is still capable of performance pre-processing directly after the image acquisition, significantly eliminating expensive raw image data movement. In a modern HMD system, the distribution of these stacked intelligent image sensors and the central AR/VR processor form the hardware backbone to realize complex CV/ML functions on device.

Within such systems, it's natural to split the machine learning workload for an application between the sensors and the centralized computer (aggregator). As is shown in~\Cref{fig:compute_partition}, for an application that requires multiple layers of convolutional neural networks and fusion of multiple input sources, the early layers can be distributed to the on-sensor processing. The feature fusion and rest of processing are on the aggregator. This way, overall system latency can be improved by leveraging direct parallel processing on sensors and reduced sensor-aggregator communication.

The success of distributed computing for DNNs between sensors and aggregator heavily relies on the network architecture to satisfy the application and hardware system constraints such as memory, communication bandwidth, latency target and etc.. 
% To achieve a collaborative inference between multiple compute modelities, p
Prior work~\cite{jeong2018computation,kangNeurosurgeonCollaborativeIntelligence2017a,eshratifar2019jointdnn,pagliari2020crime} searches network partitions (\ie, the {\mysp}s) for existing models in either exhaustive or heuristic manners. Some work~\cite{eshratifar2019bottlenet,shao2020bottlenet++,matsubara2020head,assine2021single,sbai2021cut} manually injects a bottleneck module into the model to reduce communication. However, these methods sometimes obtain naive splitting results (\ie, splitting after the last layer)~\cite{kangNeurosurgeonCollaborativeIntelligence2017a} and lead to performance degradation~\cite{matsubara2019distilled}.

The challenge of splitting DNNs comes from the complicated mutual-impact of many model and hardware factors. For example, existing (hand-crafted and searched) model architectures are designed without considering distributed computing over multiple computing modelities, and thus not suitable for splitting in the first place. In addition, the position of the splitting point as well as the inserted compression module will simultaneously impact model performance, computation offload, communication, and hardware utilization in different directions~\cite{matsubara2021split}. Heuristic and rule-based methods are limited in this context.  
% The challenge of splitting a deep model comes from the complicated entanglement of many model and hardware factors. For example, existing (hand-crafted and searched) model architectures are designed without giving consideration to distributed computing and, thus not suitable for splitting in the first place. In addition, the position of splitting point as well as inserted compression module will simultaneously impact model performance, computation offload, communication, and hardware utilization in different directions. Heuristic and rule-based methods are limited in this context.  

In this work, we adapt the neural architecture search (NAS) approach to automatically search split-aware model architectures targeting the distributed computing systems on AR/VR glasses. We propose the \method framework that jointly optimizes the task performance and system efficiency subject to resource constraints of the mobile AR/VR system featuring smart sensors. We specifically answer the following two questions:
\begin{enumerate}[parsep=0pt]
    \item Can we jointly search for optimal network architectures and their network splitting solution between sensors and the aggregator while satisfying resource constraints imposed by the underlining hardware?
    \item Can we learn an optimal network architecture to compress and fuse features from multiple sensors to the aggregator and achieve SOTA performance and efficiency compared with conventional centralized models?
\end{enumerate}

We design \method, a split-aware NAS framework, for efficient and flexible searching of the splitting module in the network in the distributed computing context where the splitting module is able to split the network, compress features along the channel dimension for communication saving as well as view fusion for multi-view tasks.

We introduce a series of techniques for module initialization and sampling to stabilize the training with information bottlenecks and mitigate introduced accuracy degradation. We further extend \method to support searching of splitting modules with view fusion for multi-view tasks. To our best knowledge, it is the first framework supporting the position searching of information compression / fusion in a multi-input neural network. Overall, our contributions are summarized as follows:
\begin{itemize}
\itemsep -3pt
    \item We propose \method, a split-aware NAS framework, for efficient and flexible position searching of splitting modules for single / multi-view task. 
    \item We introduce splitting modules for single- and multi- view tasks which can achieve model splitting, feature compression, as well as view fusion simultaneously. To search the optimal position of the splitting module, we propose to use separate supernets for sensors and the aggregator respectively, and stitch them together to form the split-aware model of interest, using the shared splitting module as the joint point. In addition, we combine the compression- / recovery- / fusion- based splitting module design with custom weights initialization, and a novel candidate networks sampling strategy to mitigate the accuracy drop due to model partition.  
    % We introduce hourglass-like compression module with an improved candidate networks sampling strategy and weight initialization to reduce communication with minimized training instability and performance drop.
    \item We evaluate \method with single-view classification and multi-view 3D classification. Empirical observations validate the importance of joint model and splitting position search to improve  both task and system performance. Our results show that optimized network architectures and model partitions discovered by \method significantly outperform existing solutions and fit the distributed computing system well on AR/VR glasses. 
\end{itemize}

\section{Background and Related Work}
We consider a system that consists of three kinds of components, $V$ sensors, one aggregator, and communication interfaces between any of sensors and the aggregator. For simplicity, we assume that sensors are homogeneous. 

A DNN is a composition of layers / computation which can be distributed to sensors and aggregator. Compared with on-sensor computing (\textit{All-on-sensor}) and conventional mobile computing (\textit{All-on-aggregator}), distributed computing has the flexibility to achieve a better balance between computation and communication, and enable high utilization of hardware, as summarized in~\Cref{tab:compare_paradigm}.

A challenge is to determine a \mysp in the DNN. 
The on-sensor (on-sen.) part $f_\texttt{sen}$, which is composed of all layers of the DNN before the \mysp, is executed on sensors' processors.
The feature $\rvz$ generated by each sensor will be uploaded to the aggregator. 
The on-aggregator (on-agg.) part $f_\texttt{agg}$, which consists of all remaining layers of the DNN after the \mysp, receives $\rvz$ and finishes its computation on the computationally more capable aggregator processor. 

The problem of finding the optimal \mysp can be summarized as follows,
% \begin{align}
\begin{equation}\label{equ:objective}
\begin{split}
    \min_{f_\texttt{sen},f_\texttt{agg}} T_\texttt{sen}(f_\texttt{sen}, \rvx) + T_\texttt{comm}(\rvz) &+ T_\texttt{agg}(f_\texttt{agg}, \rvz) \\
    \text{where}\quad\rvz = f_\texttt{sen}(\rvx)\\
    \text{s.t.}\quad \mathcal{L}(f_\texttt{agg}\circ f_\texttt{sen};\mathcal{D}^\texttt{val}) &\leq \mathrm{Loss\ \ target}\\
    \operatorname{PeakMem}(f_\texttt{sen}, \rvx) &\leq \operatorname{Mem}_\texttt{sen},\\
    % \operatorname{Bandwidth}(\rvz) \leq C
    % \label{}
\end{split}
\end{equation}
% \end{align}
% \xin{define AB here}
where $T(\cdot,~\cdot)$ is latency measurement function and $\mathcal{L}$ is the loss function, e.g., cross entropy. $\operatorname{PeakMem}(\cdot,~\cdot)$ measures the peak memory consumption of the on-sensor part, and $\operatorname{Mem}_\texttt{sen}$ is the memory size of a sensor processor.
% After splitting, the latency or energy ...

To find the optimal \mysp, one has to take several factors into consideration simultaneously as summarized in~\Cref{tab:system_factors}. Existing solutions only consider a fraction of these factors via heuristic or exhaustive approaches~\cite{jeong2018computation,kangNeurosurgeonCollaborativeIntelligence2017a,eshratifar2019jointdnn,pagliari2020crime}. DNNs typically have tens or even hundreds of layers. Therefore, exhaustive searching with retraining is costly. Some literature proposes injecting a hand-crafted compression module to reduce the feature communication~\cite{eshratifar2019bottlenet,shao2020bottlenet++,matsubara2020head,assine2021single,sbai2021cut}. 
However, manually designing and inserting the compression module not only lead to high engineering effort and training cost, but also result in sub-optimal system performance and accuracy degradation.

In addition, previous methods all consider to split existing models like VGG~\cite{Simonyan15VGG} and ResNet~\cite{he2016deep}. However, those models may not be suitable for splitting. For example, Kang~\etal~\cite{kangNeurosurgeonCollaborativeIntelligence2017a} profiles seven models and finds that their best {\mysp}s typically fall in the first or the last layers. Furthermore, injecting compression module to existing models usually leads to significant accuracy drop~\cite{matsubara2020head}.

Other work proposes dedicated lossy (or lossless) compression techniques to reduce intermediate activation and save communication, including compressive sensing~\cite{yao2020deep}, pruning~\cite{jankowski2020joint} and quantization~\cite{li2018auto,choi2018deep,cohen2020lightweight}. \method are orthogonal to, and can benefit from these prior approaches.

\section{Split-Aware NAS}
\label{sec:split-aware-nas}
We adapt NAS to solve the constrained optimization in~\Cref{equ:objective}. The goal is to minimize the systemic latency $T_\texttt{sen} + T_\texttt{comm} + T_\texttt{agg}$, and improve task performance while satisfying the hardware constraints.

We search the whole model $f_\texttt{sen}\circ f_\texttt{agg}$ and the position of splitting point, denoted as \textbf{S}plit-\textbf{A}ware- NAS (or SA-NAS). This way, all layers before (or after) the splitting point will compose $f_\texttt{sen}$ or $f_\texttt{agg}$, respectively. 
In addition, we also consider a splitting module which is inserted at the position of splitting point. The feature compression in the splitting module can reduce communication cost by reducing the channel size $c$ of feature tensor $\rvz \in \mathbb{R}^{c\times s\times s}$, but could affect training stability and result in loss of task performance. We propose several approaches to mitigate the issue.
The splitting module evolves along with $f_\texttt{sen}$ and $f_\texttt{agg}$ during SA-NAS as elaborated in~\Cref{sec:cm_arch}. Eventually, the searched network must have \textbf{exactly one} splitting module.
%and the splitting point falls in the middle of compression module.   

\fakeparagraph{Preliminary: Two-stage NAS}
Previous NAS approaches that leverage evolutionary search~\cite{real2017large,real2019regularized,suganuma2018exploiting,xie2017genetic} or reinforcement learning~\cite{tan2019mnas,zhong2018practical,zoph2016neural,zoph2018learning} require excessive training due to the large number of models to be trained in a single experiment. Recent advancement of NAS decouples the model training and architecture search into two separate stages~\cite{yu2020bignas,guo2020single,chu2021fairnas,cai2020once,hanruiwang2020hat} that significantly reduces the training cost. 
In a two-stage NAS framework, the model training stage optimizes a supernet and its all sub-networks with weight-sharing. In the subsequent architecture optimization stage, sub-networks are searched based on system constraints to yield the best task and system performance trade-off.  
Due to the above advantages, we construct our SA-NAS based on two-stage NAS framework.
\subsection{Building and Training Supernet (Stage 1)}

\fakeparagraph{Backbone of Supernet}
We  build  the  supernet’s  search  space  following FBNet-V3~\cite{dai2020fbnetv3} which uses the MobileNetv2 block~(denoted as MB, \textit{a.k.a.}, inverted residual block)~\cite{SandlerHZZC18mobilenetv2} as meta architecture. Check \Cref{appendix:build_supernet} for more details. 
We augment the supernet with SA-NAS specific search space (\Cref{fig:SANAS}) for our experiments as discussed in the following two sections.

\vspace{-0.5em}
\subsubsection{SA-NAS for Single-View SplitNets}
\begin{figure}
    \vspace{-1em}
    \centering
    \includegraphics[width=\linewidth]{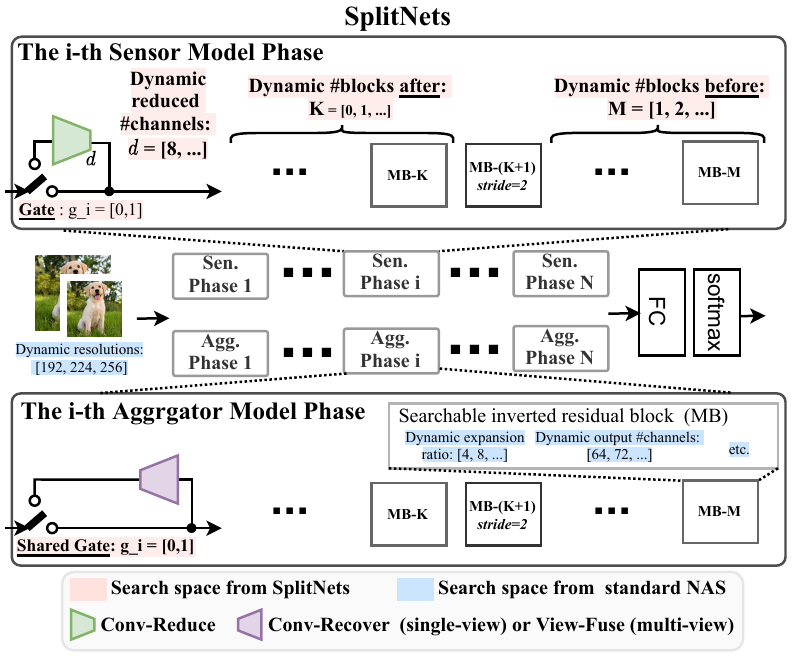}
    \caption{Architecture sampling space in training SA-NAS. Both on-sen. and on-agg. networks consist of multiple phases with each containing a split module, flexible blocks before stride, a stride-2 block, and flexible blocks after stride. The splitting module can be a reduce operator (on-sen., green) or a recover / fuse operator (on-agg., purple). SA-NAS also includes standard NAS search space (blue) such as input resolution, channel widths, and etc.}
    \label{fig:SANAS}
    % \vspace{-1.5em}
\end{figure}

Now we elaborate how to search SplitNets for a single-view task~(\eg, ImageNet classification), on searching the position of a splitting module, as shown in~\Cref{fig:SANAS}.

\begin{figure}
    \vspace{-0.5em}
    \centering
    \includegraphics[width=\linewidth]{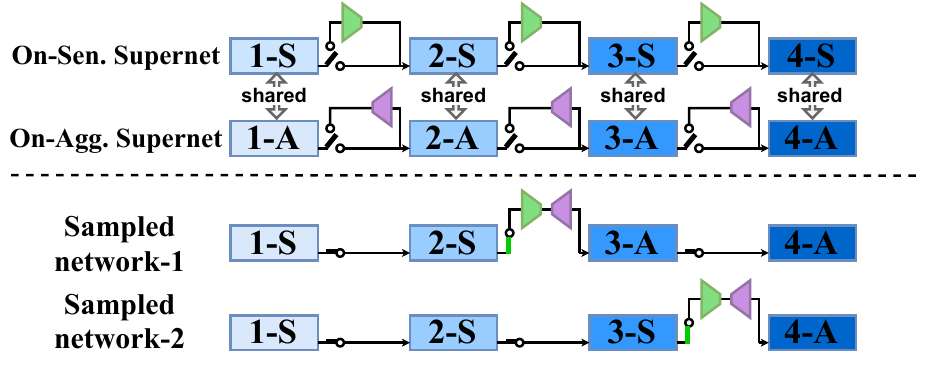}
    \caption{SA-NAS for single-view \method. At each training step, several sub-networks are sampled from the search space. A sub-network is specified by a set of choices including on-sen. blocks, splitting position, on-agg. blocks and blocks' configurations. For single-view, weight-sharing between on-sen. and on-agg. blocks are enabled since they are learning similar features.
    % can be leveraged to improve training performance.
    }
    \label{fig:singleview_stitch}
    \vspace{-3mm}
\end{figure}

The splitting module for single-view tasks consists of two parts, \textit{Conv-Reduce} and \textit{Conv-Recover}, which will be discussed in ~\Cref{sec:cm_arch} for their architectures and initialization strategy.
The splitting module could be inserted after any of inverted residual block (MB) in a supernet theoretically. However, instead of equipping every MB with a candidate splitting module, we divide a supernet into several phases and insert one splitting module for each phase to reduce the number of splitting positions to be optimized.  

Suppose a supernet has $N$ stride-2 MB blocks, We divide the whole model into $N$ phases where each phase has one stride-2 MB block. Within a phase, MB blocks before the stride-2 block have the same spatial size and similar channel sizes. We then insert one \textit{Conv-Reduce} ({green trapezium},~\Cref{fig:SANAS}) layer $\phi: \mathbb{R}^{c\times s\times s}\rightarrow\mathbb{R}^{d\times s\times s}$ in each phase for the on-sensor supernet, associated with a gate variable $\ervg_i\in\{0,~1\}$ indicating whether this layer is selected. A corresponding \textit{Conv-Recover} ({purple trapezium},~\Cref{fig:SANAS}) layer $\phi': \mathbb{R}^{d\times s\times s}\rightarrow\mathbb{R}^{c\times s\times s}$ will be inserted into the on-aggregator supernet.
% to receive on-sensor network output. 
\textit{Conv-Recover} layer shares the same gate variable with \textit{Conv-Reduce}, meaning that they are always selected, or not, at the same time. Using \textit{Conv-Recover} allows weight-sharing of a block between different sampled sub-networks specified by different \textit{Conv-Reduce} selections during training.
%can avoid abrupt change of its adjacent block's input channel size introduced by \textit{Conv-Reduce} selection. 

The number of MB blocks before (variable $M$) and after (variable $K$) the splitting module can be adjusted freely and searched through SA-NAS. This provides us with full search space of the splitting point at fine granularity, as well as a small amount of candidate networks to train. In particular, on-sen. and on-agg. blocks at the same depth learn similar features and can share weights for single-view problems. 

% \xin{we need to explain more clear why ``small amount of candidate networks''}

In sub-network sampling, once a splitting module (\ie, $g_i=1$) is selected, we take blocks before splitting from the on-sensor supernet and blocks after splitting from the on-aggregator supernet, and stitch them together using splitting module as the joint point to form a sampled network as illustrated in~\Cref{fig:singleview_stitch}.

Eventually, we need exactly one splitting module. Therefore, gate vector $\rvg=\left(\ervg_1, \dots, \ervg_M\right)$ is restricted to a one-hot vector in the resource-constrained architecture search stage.
However, during supernet training, we are free to insert more splitting modules to help reduce the maximum performance loss (see ~\Cref{appendix:training_supernet}).

\begin{figure}[!t]
    \centering
    \includegraphics[width=0.9\linewidth]{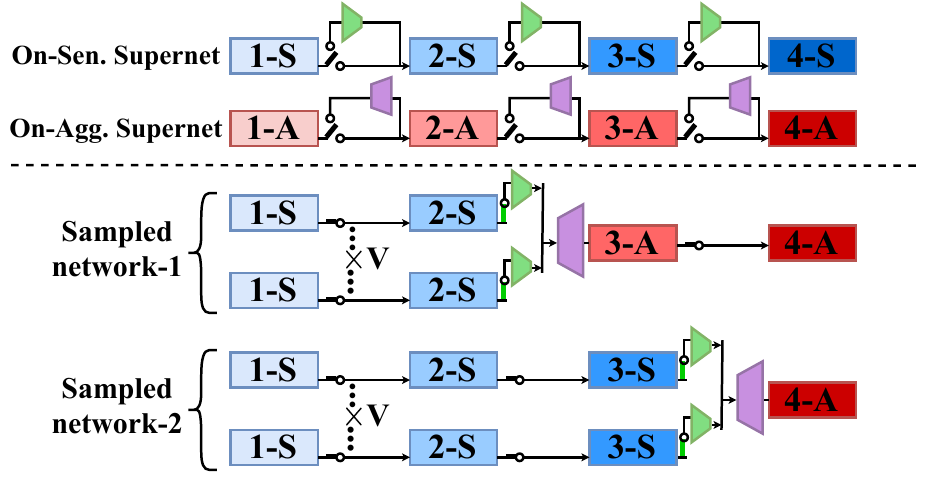}
    \caption{SA-NAS for multi-view \method. A convolution layer (\textit{Conv-Reduce}, purple trapezium) to reduce input's channel size on the sensor side. The reduced input from $V$ views will be concatenated together (\textit{View-Fuse}, green trapezium) on the aggregator. In this example, 3-A and 3-S are learning features for local view and mixed views respectively, thus their weights cannot be shared. }
    \label{fig:fusion_illu}
    \vspace{-3mm}
\end{figure}

% \vspace{-0.5em}
\subsubsection{SA-NAS for Multi-View SplitNets}

With $V$ sensors in the system, $V$ images are captured from different perspectives. Each sensor processes its captured image with the on-sen. model. Afterwards, a fusion module on the aggregator will fuse the compressed $V$ features from sensors and the on-agg. model further transforms the fused features to the final result.

Similar to searching the position of splitting module~(\ie, \textit{Conv-Reduce} and \textit{Conv-Recover}) for single-view tasks, we now use SA-NAS to search the position of splitting module~(\ie, \textit{Conv-Reduce} and \textit{View-Fuse}) for multi-view tasks. Compared with single-view case, we keep the \textit{Conv-Reduce} layers in the on-sen. supernet, but replace the \textit{Conv-Recover} layers in the on-agg. supernet with \textit{View-Fuse} layers to aggregate features across multiple views as illustrated in~\Cref{fig:fusion_illu}.

% In supernet training, we also sample $N_{sen}$ blocks from the on-sen supernet and $N_{agg}$ blocks from the on-agg supernet to form a candidate network.\xin{I cannot follow the above sentence.} 
Different from single-view case, no weight sharing can be performed between the on-sen. and on-agg. supernets because on-agg. network learns to process fused representation which contains mixed information from $V$ views, while on-sen. supernet processes only local information. In other words, blocks on-sen. and on-agg. have radically different functionalities, and it is sub-optimal to enforce a shared block to play these two roles simultaneously.

\vspace{-0.5em}
\subsubsection{Architecture of Splitting Module}
\label{sec:cm_arch}
\vspace{-0.5em}

\fakeparagraph{Single-View: Reduce and Recover} 
Inspired by ~\cite{eshratifar2019bottlenet,shao2020bottlenet++,matsubara2020head,assine2021single,sbai2021cut}, we adopt a straight-forward implementation for the splitting module in SA-NAS. As illustrated in~\Cref{fig:SANAS}, we use two convolution layers to compress features: $\phi: \mathbb{R}^{c\times s\times s} \rightarrow \mathbb{R}^{d\times s\times s}$ and recover features: $\phi': \mathbb{R}^{d\times s\times s }\rightarrow \mathbb{R}^{c\times s\times s}$ on channel dimension, where $d$ can be searched by SA-NAS. In SA-NAS, $\phi$ and $\phi'$ are always selected simultaneously.

\fakeparagraph{Multi-View: Reduce and Fuse} 
Prior work has proposed intricate fusion architecture like recurrent and graph neural networks targeting different applications~\cite{su2018deeper,ma2018learning,han2018seqviews2seqlabels,feng2018gvcnn}.

Similar to the single-view, multi-view splitting module uses a convolution layer $\phi: \mathbb{R}^{c\times s\times s} \rightarrow \mathbb{R}^{d\times s\times s}$ for communication saving, and then fuses these reduced features from $V$ views together with the \textit{View-Fuse} layer. 
Without loss of generality, we use a highly simplified fusion operation - concatenation~(See~\Cref{appendix:concat}).  \textit{View-Fuse} will concatenate $V$ views together along the channel dimension $\texttt{concat}\left(V\times \mathbb{R}^{d\times s\times s}\right) \rightarrow \mathbb{R}^{V\cdot d \times s\times s}$.

Despite of the extreme simplicity of fusion architecture, we empirically show that searching the splitting~(\ie, fusion) position jointly with network architecture using SA-NAS achieves considerable accuracy improvement, compared with SOTA approaches with dedicated fusion layers (see~\Cref{sec:results}). 

% \begin{table}[ht]
%     \centering
%     \begin{tabular}{c|c|c}
%         Method & Weight Distribution & Average \\
%         \midrule
%         Xavier~\cite{glorot2010understanding} & $\mathcal{N}\left(0, {\frac{2}{k^2\cdot{{(c_\mathrm{in}+c_\mathrm{out})/2}}}}\right)$ & Arithmetic\\
%         \midrule
%         \begin{tabular}[c]{@{}c@{}}Kaiming - Fan In \\ or -Fan Out~\cite{he2015delving}\end{tabular}  & \begin{tabular}[c]{@{}c@{}}$\mathcal{N}\left(0, {\frac{2}{k^2\cdot{{c_\mathrm{in}}}}}\right)$ \\ or $\mathcal{N}\left(0, {\frac{2}{k^2{\boldsymbol{c_\mathrm{out}}}}}\right)$\end{tabular} & No \\
%         \midrule
%         Ours & $\mathcal{N}\left(0, {\frac{2}{k^2{{\sqrt{c_\mathrm{in}\cdot c_\mathrm{out}}}}}}\right)$ & Geometric\\
%     \end{tabular}
%     \caption{Comparison of different initialization approaches. Equations in this table assume the activation function as ReLU. }
%     \label{tab:init_compare}
% \end{table}

\begin{table}[!t]
    \centering
    \begin{tabular}{ccc}
    \toprule
        {\footnotesize Initialization Method} & {\footnotesize Weights' Variance} & {\footnotesize Average} \\
        \midrule

        \begin{tabular}[c]{@{}c@{}}{\footnotesize Kaiming Fan-In or -Out}~\cite{he2015delving}\end{tabular}  &
        \begin{tabular}[c]{@{}c@{}}${\frac{2}{k^2\cdot{{c_\mathrm{in}}}}}$ or $ {\frac{2}{k^2{{c_\mathrm{out}}}}}$\end{tabular} & {\footnotesize No} \\
        % \midrule
         {\footnotesize Xavier}~\cite{glorot2010understanding} & $ {\frac{2}{k^2\cdot{{(c_\mathrm{in}+c_\mathrm{out})/2}}}}$ & {\footnotesize Arithmetic}\\
        % \midrule
        {\footnotesize Ours} & $ {\frac{2}{k^2{{\sqrt{c_\mathrm{in}\cdot c_\mathrm{out}}}}}}$ & {\footnotesize Geometric}\\
    \bottomrule
    \end{tabular}
    \caption{Comparison of different initialization approaches. Equations in this table assume the activation function as ReLU. }
    \vspace{-3mm}
    \label{tab:init_compare}
\end{table}

\fakeparagraph{Split-aware Weights Initialization} As is also observed by ~\cite{matsubara2020head}, adding compression / fusion modules into supernet training process degrades the training stability and yields suboptimal training performance. We find that the issue is caused by the gradient of the compression module.
In practice, the gradients of the compression layer are $>$10$\times$ bigger than regular layers, starting from initialization.

Take Kaiming initialization as an example~\cite{he2015delving}.
% The idea of initialization strategies
Its Fan-In mode
is to ensure that the output of each layer has zero mean and unit variance 
% (\Cref{equ:kaiming_forward})
, so the output magnitude will not explode in the forward pass. 
Similarly, the backward pass of a convolution layer is also a convolution but with the transposed weight matrix $\rmW^T$,
% (\Cref{equ:kaiming_backward})
and applying the same idea will lead to Fan-Out mode
.

Either Fan In or Fan Out works well for standard neural architectures because most of layers' input channel size $c_\mathrm{in}$ and output channel size $c_\mathrm{out}$ are similar~(see \Cref{tab:init_compare}).
% $0.5<\frac{c_\mathrm{in}}{c_\mathrm{out}}<2$. 
However, for \textit{Conv-Reduce} and \textit{Conv-Recover}, $c_\mathrm{in}$ and $c_\mathrm{out}$ are very different, thus Kaiming and Xavier cannot reconcile both forward and backward at the same time. 

In order to mitigate this conflict, we use a new initialization strategy by replacing $c_\mathrm{in}$ or $c_\mathrm{out}$ with their geometic average $\sqrt{c_\mathrm{in}\cdot c_\mathrm{out}}$. Using geometric average allows us to find a better compromise between forward and backward passes and significantly improve training stability and final accuracy as shown in~\Cref{appendix:init_numbers}.

\subsection{Resource-constrained searching (Stage 2)}
\label{sec:stage-two}
After the supernet training stage, all candidate networks are fully optimized. The next step is to search the optimal network which satisfies system constraints. Since two-stage NAS decouples the training and searching, the system constraints are fully configurable. In addition, one does not have to redo the costly supernet training when the system constraints change. 

Since the searching space of supernet is extremely huge, evolution algorithm is adopted to accelerate the searching process~\cite{yu2020bignas,guo2020single,chu2021fairnas,cai2020once,hanruiwang2020hat}. Specifically, we mutate and crossover pairs of current generation's networks to generate their children networks. All children networks are evaluated and selected 
according to their hardware and task performance. The selected children networks will be used to populate candidate networks for the next generation. After several generations, optimal networks can be found. 

We consider two kinds of hardware constraints simultaneously. 
% as marked in~\Cref{tab:system_factors}. 
1) Soft constraints, such as overall latency under the same task accuracy: Children networks with top-$k$ smallest latency's will be selected. 2) Hard constraints, such as peak memory usage: Regardless of accuracy and performance, only children networks that satisfy all hard constraints can be selected to the next generation. 

\begin{figure}[t]
    \centering
    \includegraphics[width=0.8\linewidth]{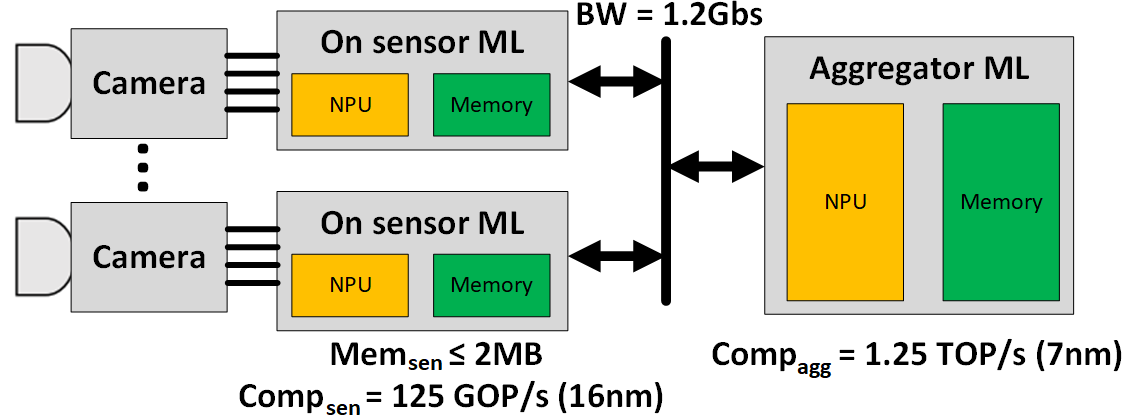}
    \vspace{-1mm}
    \caption{Hardware model for the target system. On-sensor compute uses a less advanced technology node (16nm) with limited on-chip memory. Smart sensors are connected to an aggregator using a shared 1.2 Gb/s bus. Aggregator uses advanced process (7nm) and has faster compute with larger memory.}
    \label{fig:HW_model}
    \vspace{-3mm}
\end{figure}

\fakeparagraph{System Hardware Modeling}
Given a sampled network,
its hardware performance is modeled through a hardware simulator. In this work, we use a hardware simulator customized for a realistic HMD system~\cite{qin2021virtual,zheng2020four} with smart sensors (see \Cref{fig:HW_model}). The on-sensor processor is equipped with a 16nm neural processing unit (NPU) with peak performance of $\operatorname{Comp}_\texttt{sen}=125$ GOP/s and peak memory of $\operatorname{Mem}_\texttt{sen}=2$ MB. The aggregator processor is modeled with a powerful 7nm NPU with peak performance of $\operatorname{Comp}_\texttt{agg}=1.25$ TOP/s and sufficient on-chip memory. The communication between the sensors and the aggregator is modeled with a high-performance shared bus for HMD with peak bandwidth of $\operatorname{BW}_\texttt{comm}=1.2$ Gbs. The high speed bus will be shared between multiple sensors in practice. More implementation details are summarized in the appendix.

\section{Results}
\label{sec:results}

We validate that \method can satisfy all system hardware constraints and find an optimized model with competitive accuracy and the best system performance compared with SOTA models crafted by hands or searched by standard NAS methods. 
Specifically, we evaluate our \method for single-view system on ImageNet classification~(\Cref{sec:result:imagenet})  and \method for multi-view system on 3D classification~(\Cref{sec:result:modelnet}).

\begin{figure}[t]
    \centering
    \includegraphics[width=\linewidth]{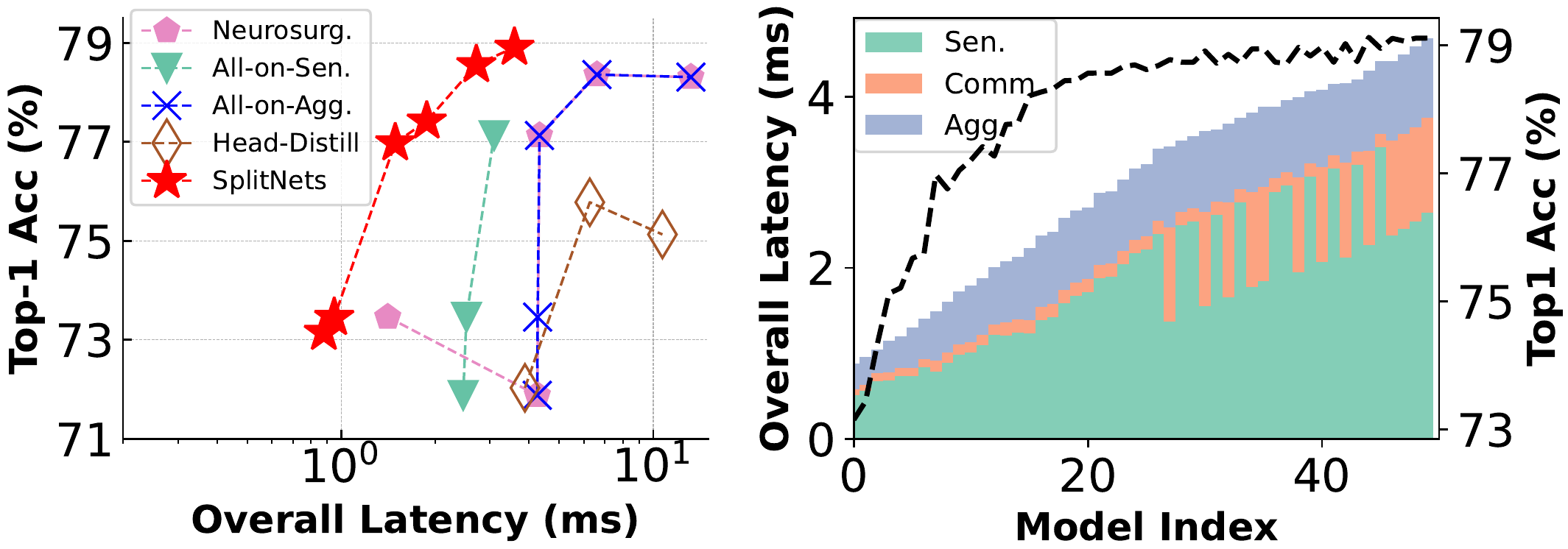}
    \vspace{-4mm}
    \caption{\textbf{Left}: Comparison of \method with prior approaches using different network architectures on ImageNet from~\Cref{tab:imagenet_main}. \textbf{Right}: Visualization of \method searched models' accuracy (dashed line and the right y-axis) and overall latency breakdown (the stacking bar plot and the left y-axis). Each searched model's overall latency is represented by a vertical stacked bar. From bottom to top of each bar, the length's of three colors represent $T_\texttt{sen},~T_\texttt{comm},~T_\texttt{agg}$ respectively.}
    \label{fig:imagenet_acc}
    \vspace{-0.8em}
\end{figure}

\begin{table*}[ht]
\centering
\resizebox{\linewidth}{!}{
\begin{tabular}{c|ccc||cccc|cc|ccc|c}
\toprule
    \multirow{2}{*}{\begin{tabular}[c]{@{}c@{}} HW Constraints \end{tabular}} &
    \multirow{2}{*}{\begin{tabular}[c]{@{}c@{}} Method \end{tabular}} &
    \multirow{2}{*}{\begin{tabular}[c]{@{}c@{}} Backbone \end{tabular}} &
    \multirow{2}{*}{\begin{tabular}[c]{@{}c@{}} Top-1 \end{tabular}} &
    \multicolumn{4}{c|}{On-sensor} &
    \multicolumn{2}{c|}{\begin{tabular}[c]{@{}c@{}} Comm. \end{tabular}} &
    \multicolumn{3}{c|}{\begin{tabular}[c]{@{}c@{}} On-aggregator \end{tabular}} &    
    \multirow{2}{*}{\begin{tabular}[c]{@{}c@{}} Overall \\ Latency \end{tabular}}
    \\
    % \cmidrule(lr){5-8}  \cmidrule(lr){9-10} \cmidrule(lr){11-13} 
    % \cline{5-13}
    &&&& \#Params & \#OPs& Peak & Latency & Size & Latency & \#Params & \#OPs& Latency &   \\
    
\midrule
%%%%%%--------------- HW Constraints -----------
\multirow{23}{*}{\begin{tabular}[c]{@{}c@{}} 
\textbf{$\operatorname{Mem}_\texttt{sen}$ $\leq$ 2MB} \\
\\
\textbf{$\operatorname{Comp}_\texttt{sen}$ (16nm) =}\\
\textbf{$\mathbf{125}$ GOP/s}\\
\\
\textbf{$\operatorname{Comp}_\texttt{agg}$ (7nm) = }\\
\textbf{$\mathbf{1.25}$ TOP/s}\\
\\
\textbf{$\operatorname{BW}_\texttt{comm}$ = }\\
\textbf{37.5 MB/s}\\
\\
\\
\textbf{$*$: For all models,}\\
\textbf{We assume weights} \\
\textbf{and activation} \\
\textbf{are quantized} \\
\textbf{to 8-bit without}\\
\textbf{loss of accuracy.} \\
% \textbf{}  
\end{tabular}} 
%%%%%%--------------- HW Constraints -----------
&
\multirow{5}{*}{\begin{tabular}[c]{@{}c@{}} All-on-sen.  \end{tabular}} &
MobileNet-v2 & 71.88 & 3.51M & 301M & \textcolor{mypink}{5.90MB} & 2.46ms & 0 & 0 & 0 & 0 & 0 & 2.46ms \\
&& MNASNet-1.0$^{\boldsymbol{\mathsection}}$ & 73.46 & 4.38M & 314M & \textcolor{mypink}{5.59MB} & 2.52ms & 0 & 0 & 0 & 0 & 0 & 2.52ms\\
& & EfficientNet-B0$^{\boldsymbol{\mathsection}}$ & 77.13 & 5.30M & 386M & \textcolor{mypink}{7.70MB} & 3.09ms & 0 & 0 & 0 & 0 & 0 & 3.09ms\\
&& ResNet-152 & 78.31 & 60.2M & 11.5G & \textcolor{mypink}{120 MB} & 92.1ms & 0 & 0 & 0 & 0 & 0 & 92.1ms  \\
&& RegNetX-3.2GF$^{\boldsymbol{\mathsection}}$ & 78.36 & 15.3M & 3.18G & \textcolor{mypink}{29.3MB} & 25.4ms & 0 & 0 & 0 & 0 & 0 & 25.4ms \\
\cline{2-14}
 &
 %%%%################## Agg ############
\multirow{5}{*}{\begin{tabular}[c]{@{}c@{}} All-on-agg.  \end{tabular}} &
MobileNet-v2 & 71.88 & 0 & 0 & \textcolor{mygreen}{0} & 0 & $3\cdot224^2$ & 4.01ms & 3.51M & 301M & 0.24ms & 4.25ms\\
&& MNASNet-1.0$^{\boldsymbol{\mathsection}}$ & 73.46 & 0 & 0 & \textcolor{mygreen}{0} & 0 & $3\cdot224^2$ & 4.01ms &  4.38M & 314M & 0.25ms & 4.26ms\\
&& EfficientNet-B0$^{\boldsymbol{\mathsection}}$ & 77.13 & 0 & 0 & \textcolor{mygreen}{0} & 0 & $3\cdot224^2$ & 4.01ms & 5.3M & 386M & 0.31ms & 4.32ms\\
&& ResNet-152 & 78.31 & 0 & 0  & \textcolor{mygreen}{0} & 0 & $3\cdot224^2$ & 4.01ms & 60.2M & 11.5G & 9.21ms & 13.2ms \\
&& RegNetX-3.2GF$^{\boldsymbol{\mathsection}}$ & 78.36 & 0 & 0  & \textcolor{mygreen}{0} & 0 & $3\cdot224^2$ & 4.01ms & 15.3M & 3.18G & 2.54ms & 6.60ms\\
\cline{2-14}
&
%%##################### Neurosurgeon
\multirow{5}{*}{\begin{tabular}[c]{@{}c@{}} Neurosurgeon  \end{tabular}} &
MobileNet-v2 & 71.88 & 0 & 0 & \textcolor{mygreen}{0} & 0 & $3\cdot224^2$ & 4.01ms & 3.51M & 301M & 0.24ms & 4.25ms\\
&& MNASNet-1.0$^{\boldsymbol{\mathsection}}$ & 73.46 & 82.0K & 103~M & \textcolor{mygreen}{1.29MB} & 0.83ms & $80\cdot14^2$ & 0.84ms & 4.30M & 212M & 0.17ms & 1.82ms\\
% && EfficientNet-B1$^{\boldsymbol{\mathsection}}$ & 78.64 & 2K & 98.82M & 1.8MB & 0.493ms & (16,120,120) & 3.07ms & & &\\
&& EfficientNet-B0$^{\boldsymbol{\mathsection}}$ & 77.13 & 0 & 0 & \textcolor{mygreen}{0} & 0 & $3\cdot224^2$ & 4.01ms & 5.3M & 386M & 0.31ms & 4.32ms\\
&& ResNet-152 & 78.31 & 0 & 0 & \textcolor{mygreen}{0} & 0 & $3\cdot224^2$ & 4.01ms & 60.2M & 11.5G & 9.21ms & 13.2ms \\
&& RegNetX-3.2GF$^{\boldsymbol{\mathsection}}$ & 78.36 & 0 & 0 & \textcolor{mygreen}{0} & 0 & $3\cdot224^2$ & 4.01ms & 15.3M & 3.18G & 2.54ms & 6.60ms\\
\cline{2-14}
&
%%###################### Head Distill
\multirow{3}{*}{\begin{tabular}[c]{@{}c@{}} Head Distill  \end{tabular}} &
ResNet-152$^{\boldsymbol{\dagger}}$ & 75.13$\boldsymbol{\downarrow}$ & 12.8K & 125M & \textcolor{mygreen}{0.82MB} & 1.00ms & $12\cdot29^2$ & 0.27ms & 60.8M & 11.8G & 9.45ms & 10.72ms \\
&& DenseNet-169$^{\boldsymbol{\dagger}}$ & 72.03$\boldsymbol{\downarrow}$ & 12.8K & 125M & \textcolor{mygreen}{0.82MB} & 1.00ms & $12\cdot29^2$ & 0.27ms &  14.9M & 3.27G & 2.61ms & 3.88ms \\
&& Inception-v3$^{\boldsymbol{\dagger}}$ & 75.78$\boldsymbol{\downarrow}$ & 12.8K & 223M & \textcolor{mygreen}{1.45MB} & 1.78ms & $12\cdot38^2$ & 0.46ms & 24.1M & 5.03G & 4.02ms & 6.26ms\\
\cline{2-14}
&
% ##################.  ours ##############
\multirow{5}{*}{\begin{tabular}[c]{@{}c@{}} \method  \end{tabular}} &
A$^{\boldsymbol{\mathsection}}$ & 73.15 & 93.4K & 63.7M & \textcolor{mygreen}{0.68MB} & 0.51ms & $18\cdot12^2$ & 0.07ms & 6.73M & 372M & 0.30ms & 0.88ms\\
&& B$^{\boldsymbol{\mathsection}}$ & 73.45 & 0.12M & 70.5M & \textcolor{mygreen}{1.00MB} & 0.56ms & $18\cdot12^2$ & 0.07ms & 7.43M & 402M & 0.32ms & 0.95ms \\
&& C$^{\boldsymbol{\mathsection}}$ & 76.98 & 92.8K & 98.6M & \textcolor{mygreen}{1.14MB} & 0.79ms & $18\cdot16^2$ & 0.13ms & 7.63M & 727M & 0.58ms & 1.49ms \\
&& D$^{\boldsymbol{\mathsection}}$ & 77.42 & 208K & 137~M & \textcolor{mygreen}{1.52MB} & 1.10ms & $18\cdot16^2$ & 0.13ms & 7.85M & 831M & 0.66ms & 1.88ms \\
&& E$^{\boldsymbol{\mathsection}}$ & 78.56 & 211K & 214~M & \textcolor{mygreen}{1.54MB} & 1.71ms & $18\cdot18^2$ & 0.16ms & 7.85M  & 1050M & 0.84ms & 2.71ms \\
&& F$^{\boldsymbol{\mathsection}}$ & 78.91 & 98.0K & 194~M & \textcolor{mygreen}{1.76MB} & 1.55ms & $32\cdot36^2$ & 1.11ms & 8.13M  & 1.18G & 0.94ms & 3.59ms \\
\bottomrule
\end{tabular}
}
% \vspace{-0.13in}
% \begin{flushleft}
% $^{\boldsymbol{\mathsection}}$: {\small The backbone model is from NAS methods.} \quad
% $^{\boldsymbol{\dagger}}$: {\small The model, splitting point and accuracy are directly from ``Head Distill''.} 
% \end{flushleft}
\vspace{-2mm}
\caption{Overall latency comparison on single-sensor system for ImageNet classification task. Green (or red) numbers mean the on-sensor memory consumption is less (or more) than a sensor's memory constraint. $^{\boldsymbol{\mathsection}}$: {\small The backbone model is from NAS methods.} \quad
$^{\boldsymbol{\dagger}}$: {\small The model, splitting point and accuracy are directly from ``Head Distill''~\cite{matsubara2020head}}.}
\label{tab:imagenet_main}
% \vspace{-3mm}
\end{table*}

\begin{figure*}[!t]
  \begin{minipage}{.69\textwidth}
    \centering
    \resizebox{\linewidth}{!}{
    \begin{tabular}{cccccccc}
    \toprule
    \multirow{2}{*}{Method} &  \multirow{2}{*}{Backbone} & \multirow{2}{*}{\#Params} & \multirow{2}{*}{\#OPs} & \multicolumn{3}{c}{Fusion} & \multirow{2}{*}{Top-1} \\
    \cline{5-7}
    & & & & Position & Feature size & Arch. & \\
    
    \midrule
    MVCNN-su & VGG-M & 90.5M & 34.6G & Last Conv & $512\cdot13^2$ &View-Pool & 81.1  \\
    MVCNN-su & MobileNet-v3 & 2.54M & 661M & Last Conv & $576\cdot~7^2$ &View-Pool & 86.0  \\
    MVCNN-su & MNASNet-0.5 & 2.22M & 1.24G & Last Conv & $1280\cdot7^2$ & View-Pool & 91.4 \\
    MVCNN-su & EfficientNet-B0 & 5.29M & 4.62G & Last Conv & $1280\cdot7^2$ & View-Pool & 92.1 \\
    % GVCNN & GoogLeNet & E & 88.8  \\
    MVCNN-new & VGG-11 & 132~M & 90.0G & Last Conv & $512\cdot~7^2$ &View-Pool & 88.7   \\
    MV-LSTM & ResNet-18 & 11.2M & 21.9G &  Last Conv & $512\cdot~7^2$ &Bi-LSTM & 89.1   \\
    SeqViews & VGG-19 & 144M & 235~G & Penultimate FC & $4096\cdot1$ & Bi-RNN & 89.3 \\
    Auto-MVCNN & AM-c24 & 2.10M & 3.20G &  Last Conv & - & Searched & 90.5 \\
    Auto-MVCNN & AM-c36 & 4.70M & 6.90G &  Last Conv & -  & Searched & 91.0 \\
    \midrule
    \method & A & 1.39M & 513M &   6-th\ (tot.\ \ \ 8) block & $16\cdot6^2$ & Concat. & 92.3 \\
    \method & B & 2.44M & 759M &  8-th (tot. 13) block & $6\cdot12^2$ & Concat.& 93.0\\
    \method & C & 2.32M & 1.83G &  8-th (tot. 13) block & $6\cdot14^2$ & Concat. & 93.8\\
    \bottomrule
    \end{tabular}
    }
    \captionof{table}{Performance comparison on multi-view 3D classification for MVCNN-su~\cite{su2015multi}, MVCNN-new~\cite{su2018deeper}, MV-LSTM~\cite{ma2018learning},  SeqViews~\cite{han2018seqviews2seqlabels}, Auto-MVCNN~\cite{li2020auto}, and \method. We assume that all models are not pre-trained on ImageNet. 
    % Despite of the simplicity of {\method}' fusion module, \method achieve the best accuracy due to that its fusion can be learnt for the optimal position.
    }
    \label{tab:mvcnn-performance}
  \end{minipage}\quad~~
  \begin{minipage}{0.28\textwidth}
    \centering
    \includegraphics[width=0.69\linewidth]{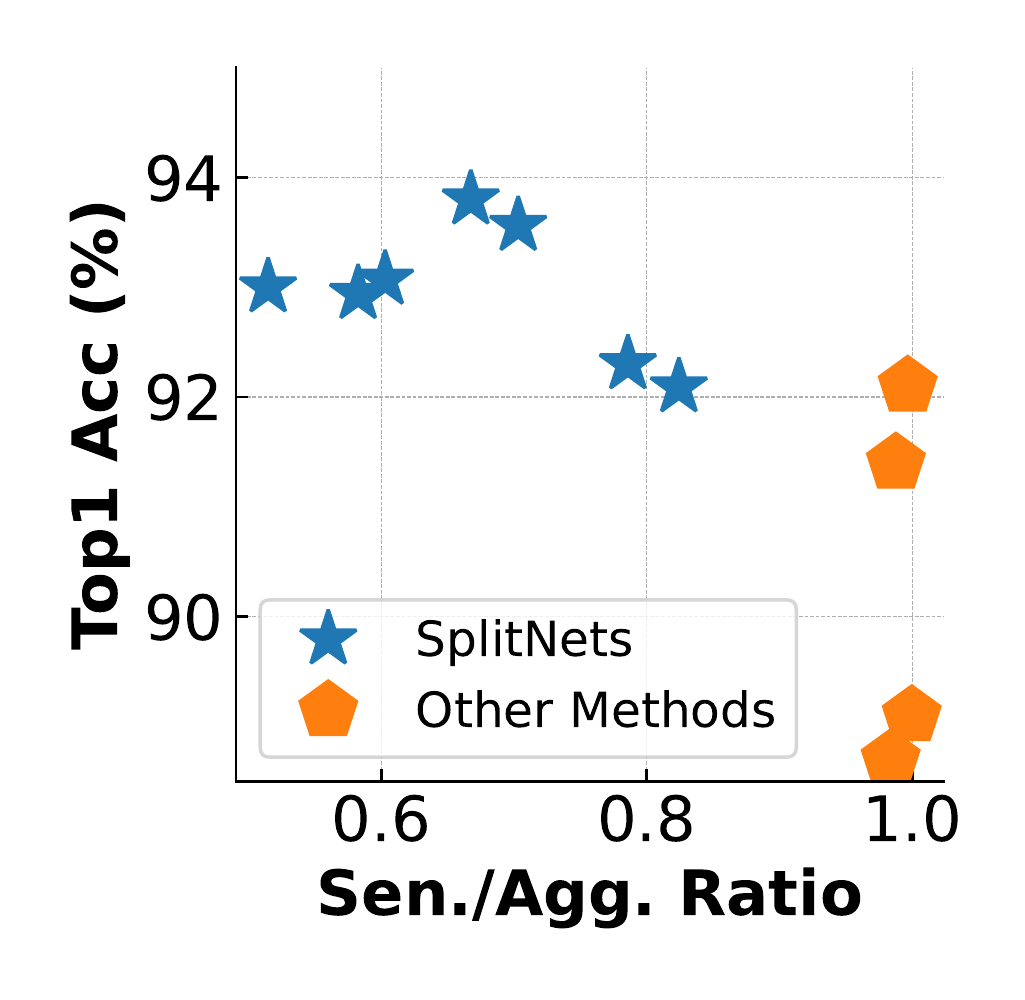}
    \vspace{-1mm}
    \caption{We compare achieved accuracy as a function of the ratio, \uline{\#OPs on one sen.} / (\uline{\#OPs on one sen.} + \uline{\#OPs on the agg.}). Since prior methods conduct fusion around the final layer, their ratios are $\sim$1. In contrast, \method are able to explore various positions and pick the best one.}
    \label{fig:mvcnn_fusion_position}
  \end{minipage}
  \vspace{-3mm}
\end{figure*}

\subsection{Single-view Task: ImageNet}
\label{sec:result:imagenet}

For the distributed computing with a single-view system, we use the public large-scale single-view classification dataset, ImageNet, to train and test the method and evaluate performance. Results are summarized in~\Cref{tab:imagenet_main}.

\begin{table*}[t]
\centering
\resizebox{\linewidth}{!}{
\begin{tabular}{c|ccc||cccc|cc|ccc|c}
\toprule
    \multirow{2}{*}{\begin{tabular}[c]{@{}c@{}} HW Constraints \end{tabular}} &
    \multirow{2}{*}{\begin{tabular}[c]{@{}c@{}} Split \end{tabular}} &
    \multirow{2}{*}{\begin{tabular}[c]{@{}c@{}} Backbone \end{tabular}} &
    \multirow{2}{*}{\begin{tabular}[c]{@{}c@{}} Top-1 \end{tabular}} &
    \multicolumn{4}{c|}{\begin{tabular}[c]{@{}c@{}} On-each-sensor \end{tabular}} &
    \multicolumn{2}{c|}{\begin{tabular}[c]{@{}c@{}} Comm. \end{tabular}} &
    \multicolumn{3}{c|}{\begin{tabular}[c]{@{}c@{}} On-aggregator \end{tabular}} &    
    \multirow{2}{*}{\begin{tabular}[c]{@{}c@{}} Overall \\ Latency \end{tabular}}
    \\
    % \cline{5-7}
    % \cmidrule(lr){5-8}  \cmidrule(lr){9-10} \cmidrule(lr){11-13}
    &&&& \#Params & \#OPs & Peak & Latency & Size & Latency & \#Params & \#OPs& Latency &   \\
    
\midrule
%%%%%%--------------- HW Constraints -----------
\multirow{13}{*}{\begin{tabular}[c]{@{}c@{}} 
\textbf{Sensor $\times$ 12}\\
\textbf{Aggregator $\times$ 1}\\
\\
\textbf{$\operatorname{BW}_\texttt{comm}$ = }\\
\textbf{12.5 MB/s}\\
\\
\textbf{$*$: The others}\\
\textbf{are same as}\\
\textbf{as~\Cref{tab:imagenet_main}.}
% \textbf{Peak Mem. $\leq$ 2MB}\\
% \\
% \textbf{Comp. Sen. (16nm)=}\\
% \textbf{$\mathbf{125}$ MMAC/s}\\
% \\
% \textbf{Comp. Agg. (7nm)= }\\
% \textbf{$\mathbf{1.25}$ GMAC/s}\\
% \\
% \textbf{Comm. Speed= }\\
% \textbf{12.5 MB/s}\\
% \\ 
% \textbf{$*$: Weights and } \\
% \textbf{activation are} \\
% \textbf{quantized to 8-bit.} \\
% \textbf{}  
\end{tabular}} 
&
%########################### On Sensor. ##############
% \multirow{5}{*}{\begin{tabular}[c]{@{}c@{}} All-on-sen.  \end{tabular}} &
% VGG-11 & 88.7 & 132~M & 90.0G &   \\
% && ResNet-18 & 89.1 \\
% && MobileNet-v3 & - \\
% && MNASNet-0.5$^{\boldsymbol{\dagger}}$ & - \\
% && EfficientNet-B0$^{\boldsymbol{\dagger}}$ & - \\
% \cline{2-14}
%  &
 %&&&&&&&&&&&&& Agg #######################
\multirow{5}{*}{\begin{tabular}[c]{@{}c@{}} All-on-agg.  \end{tabular}} &
VGG-11 & 88.7 & 0 & 0 & 0 & 0 & $3\cdot224^2$ & 12.0ms & 132~M & 90.0G & 73.2ms & 84.9ms \\
&& ResNet-18 & 89.1 & 0 & 0 & 0 & 0 & $3\cdot224^2$ & 12.0ms & 11.7M & 21.8G & 17.7ms & 29.7ms  \\
&& MobileNet-v3 & 86.0 & 0 & 0 & 0 & 0 & $3\cdot224^2$ & 12.0ms & 2.55M & 661M & 0.58ms & 12.6ms\\
&& MNASNet-0.5 $^{\boldsymbol{\dagger}}$ & 91.4 & 0 & 0 & 0 & 0 & $3\cdot224^2$ & 12.0ms & 2.22M & 1.24G & 1.01ms & 13.0ms \\
&& EfficientNet-B0 $^{\boldsymbol{\dagger}}$ & 92.1 & 0 & 0 & 0 & 0 & $3\cdot224^2$ & 12.0ms & 5.29M & 4.62G & 3.76ms & 15.8ms\\
\cline{2-14}
&
%############### split at fusion #############
\multirow{5}{*}{\begin{tabular}[c]{@{}c@{}} Split at fusion  \end{tabular}} &
VGG-11 & 88.7 & 9.22M & 7.49G & \textcolor{mypink}{12.4MB} & 59.9ms & $512\cdot7^2$& 2.00ms & 123M & 123M & 0.10ms & 62.1ms \\
&& ResNet-18 & 89.1 & 11.2M & 1.81G & \textcolor{mypink}{12.8MB} & 14.5ms & $512\cdot7^2$ & 2.00ms & 0.51M & 0.51M & 0.40ns & 16.5ms  \\
&& MobileNet-v3 & 86.0 & 0.93M & 54.9M & \textcolor{mygreen}{1.38MB} & 0.44ms & $576\cdot7^2$ & 2.26ms & 1.62M & 1.62M & 1.29ns & 2.71ms   \\
&& MNASNet-0.5$^{\boldsymbol{\mathsection}}$ & 91.4 & 0.94M & 103M & \textcolor{mygreen}{1.54MB} & 0.83ms & $1280\cdot7^2$ & 5.03ms & 1.28M & 1.28M & 1.03ns & 5.83ms \\
&& EfficientNet-B0$^{\boldsymbol{\mathsection}}$ & 92.1 & 4.01M & 385M & \textcolor{mypink}{6.42MB} & 3.08ms & $1280\cdot7^2$ & 5.03ms & 1.28M & 1.28M & 1.03ns & 8.03ms\\
\cline{2-14}
&
\multirow{3}{*}{\begin{tabular}[c]{@{}c@{}} \method  \end{tabular}} &
A$^{\boldsymbol{\mathsection}}$ & 92.3 & 0.14M & 41.8M & \textcolor{mygreen}{0.06MB} & 0.33ms & $16\cdot6^2$ & 0.05ms & 1.25M & 11.4M & 0.01ms & 0.39ms\\
&& B$^{\boldsymbol{\mathsection}}$ & 93.0 & 0.17M & 58.7M & \textcolor{mygreen}{0.08MB} & 0.47ms & $6\cdot12^2$ & 0.07ms & 2.27M & 55.4M & 0.06ms & 0.59ms  \\
&& C$^{\boldsymbol{\mathsection}}$ & 93.8 & 0.18M & 147M & \textcolor{mygreen}{0.11MB} & 1.17ms & $6\cdot14^2$ & 0.09ms & 2.15M & 73.3M & 0.07ms & 1.34ms \\
\bottomrule
\end{tabular}
}
% \vspace{+0.05in}
% \begin{flushleft}
% $^{\boldsymbol{\dagger}}$: {\small The backbone model is from NAS methods.}
% \end{flushleft}
\caption{Overall latency comparison on a multi-sensor system with a 3D classification task. Since 12 sensors are assumed in this task, each sensor is allocated 12.5MB/s bandwidth. Green (or red) numbers mean the on-sensor memory consumption is less (or more) than a sensor's memory constraint. $^{\boldsymbol{\mathsection}}$: {\small The backbone model is from NAS methods.}}
\label{tab:more_models_transfer}
\end{table*}

\fakeparagraph{\method training / searching configurations} In supernet training, we sample 5 kinds of networks (see ~\Cref{appendix:training_supernet}) and optimize for 360 epochs with batch size 4096 using the standard training receipt from \cite{wang2021attentivenas}. In resource-constrained searching, we sample 512 candidate networks with 20 generations using evolution algorithm~\cite{dai2020fbnetv3,wang2021attentivenas,hanruiwang2020hat}.

\fakeparagraph{Comparing \method against SOTA Methods} we compare \method against several existing models which are handcrafted or searched by existing NAS methods. The compared models include MobileNet-v2~\cite{SandlerHZZC18mobilenetv2}, MNASNet~\cite{tan2019mnas}, EfficientNet~\cite{tan2019efficient}, ResNet~\cite{he2016deep}, DenseNet~\cite{huang2017densely}, Inception-v3~\cite{szegedy2016rethinking}, and RegNet~\cite{radosavovic2020designing}, where the first three architectures are specifically designed for mobile on-device computing. 
To make the evaluation more realistic to real-time applications~\cite{liu2020intelligent}, we assume each of the four sensors is allocated 25\% bandwidth of the shared bus. We also assume that all models' weights and activation are quantized to 8-bit without accuracy degradation. (And literature has shown this assumption is reasonable.~\cite{haq,li2019additive,esser2019learned,fan2020training}) If models have larger bitwidth, \method will benefit more because the communication saving from feature compression is more pronounced.

We compare SA-NAS approach with four baseline model split methods:
\begin{itemize}[parsep=0pt]
    \item ``{All-on-sen.}'': All computation happens on the sensor. The communication overhead is negligible but the peak memory often exceeds sensor's memory capacity. In our experiments, we discover that all existing models, even lightweight ones, require $>$5MBs peak memory, which are too large for on-sensor deployment.
    \vspace{-0.5em}
    \item ``{All-on-agg.}'': Transmit raw image to aggregator and perform all computation on aggregator. Communication becomes the bottleneck, and the system is not scalable with increasing number of sensors.
    %with size (3,224,224) (or even larger like 299 in Inception-v3) is the latency bottleneck.
    \vspace{-0.5em}
    \item ``{Neurosurgeon~\cite{kangNeurosurgeonCollaborativeIntelligence2017a}}'': a heuristic method which profiles each layer and exhaustively searches every possible splitting position to find the optimal partition. Neurosurgeon performs the model split for many networks and finds that the optimal split often falls in the beginning or the last layer, which degenerates to ``{All-on-agg.}'' or ``{All-on-sen.}''. We apply ``Neurosurgeon~\cite{kangNeurosurgeonCollaborativeIntelligence2017a}'' on recent efficient architectures and observe similar results. This validates the necessity of joint designing network and the split.
    %, which is missed by prior work.
    %Among all the tested architectures, only MNASNet-1.0~\cite{tan2019mnas} can be split in the middle of the network.
    \vspace{-0.5em}
    \item ``{Head Distill~\cite{hinton2015distilling}}'': a method that first manually determines the position of splitting point and then replaces the head part with a smaller networks through knowledge distillation. This approach is able to significantly reduce the communication cost compared with ``All-on-agg.'' and satisfy the hardware constraints. However, ``Head Distill~\cite{hinton2015distilling}'' introduces 1.5\%-3.6\% accuracy drop even for parameter-redundant architectures like ResNet-152, DenseNet-169, and Inception-v3. Also, exhaustively searching the splitting point and hyper-parameters of the on-sen. network in ~\cite{hinton2015distilling} also requires redoing the costly training, which is inefficient for deployment. 
    %The accuracy drop comes from neither inappropriate splitting point or sub-optimal head network architecture
\end{itemize}

In the bottom of~\Cref{tab:imagenet_main}, we report a series of networks searched by SA-NAS. Under different accuracy targets, \method consistently outperforms prior methods (``Neurosurgeon''~\cite{kangNeurosurgeonCollaborativeIntelligence2017a} and ``Head Distill''~\cite{hinton2015distilling}) on overall latency while satisfying all hardware constraints.  In addition, \method are able to balance the workload of sensor and aggregator and make trade-off among communication, accuracy and computation when system hardware configuration changes (See ~\Cref{appendix:imagenet_network_change}).

\subsection{Multi-view Task: 3D Classification}
\label{sec:result:modelnet}

\begin{figure*}[!h]
    \centering
    \includegraphics[width=0.95\linewidth]{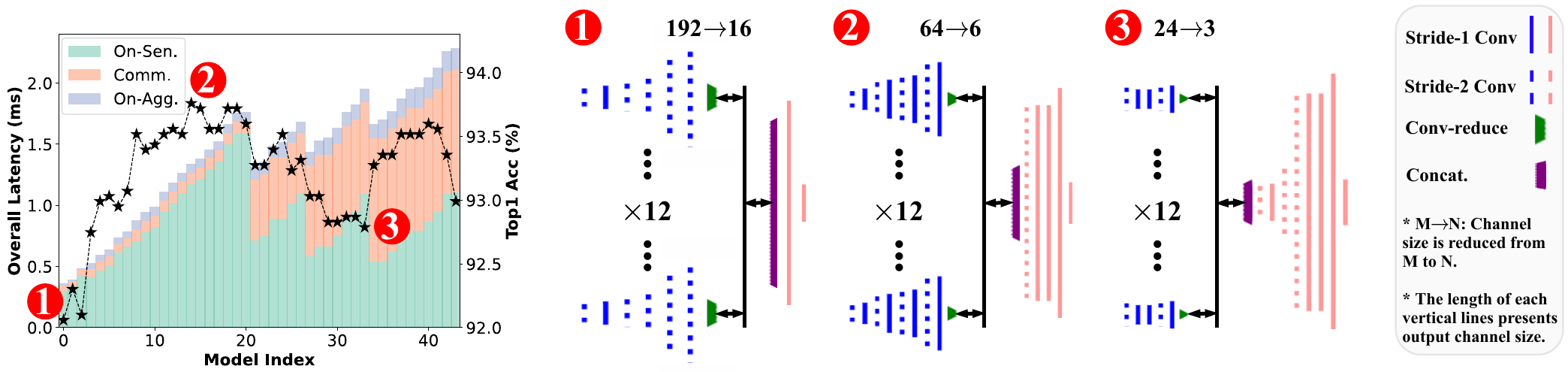}
    \vspace{-1mm}
    \caption{\textbf{Left}: Accuracy (stars and the right y-axis) and overall latency breakdown (the stacking bar plot and the left y-axis) of searched networks by \method. Each searched model's overall latency is represented by a vertical stacked bar. From bottom to top of each bar, the length's of three colors represent $T_\texttt{sen},~T_\texttt{comm},~T_\texttt{agg}$ respectively.
    \textbf{Right}: Visualization of three searched networks by \method. The length of vertical lines represents the output channel size of a convolution layer. Dash means that the stride size is two. The green and purple trapezoids are \textit{Conv-Reduce} and \textit{View-Fuse}. Among the three networks, the first and third networks conduct fusion either too `late' or too `early', leading sub-optimal task and system performance.}
    \label{fig:mvcnn_breakdown}
    \vspace{-3mm}
\end{figure*}

We also evaluate \method on a multi-view system with 3D Warehouse classification dataset that consists of 3D CAD models from 40 categories. CAD models in this benchmark come from 3D Warehouse~\cite{3dwarehouse}. Each 3D object is captured from 12 different views by 12 sensors, where each view is a 2D image of size $3\times224\times224$.

\fakeparagraph{Searching Fusion Module's Position Benefits Accuracy} In~\Cref{tab:mvcnn-performance}, we compare the task performance of \method models against existing models with various backbones~\cite{Simonyan15VGG,tan2019mnas,howard2019searching,tan2019efficient,he2016deep} with dedicated fusion modules~\cite{li2020auto}. We assume all models are trained from scratch. Prior methods do not have the flexibility in adjusting the position of fusion module, and typically fuse features after the last convolution. \method are able to explore all possible fusion positions~(\Cref{fig:mvcnn_breakdown}, Right) and learn to pick the optimal split. Despite of the simplicity of fusion used, \method significantly outperforms prior methods with fewer parameters and less computation.  
%, \ie, learn to balance computation before and after the fusion module~(\Cref{fig:mvcnn_fusion_position}). Our results reveal that 

\fakeparagraph{Superior Overall Latency with SOTA Accuracy} We further demonstrate the benefits of \method for distributed computing on a 12-sensor system in~\Cref{tab:more_models_transfer}. Compared to single-view system, multi-view system has heavier communication burden because multiple sensors have to share a single serial communication bus. As a result, the communication advantage of \method is more pronounced. 
% In addition, with the ability of fusion position searching, unnecessary layers can be reduced to achieve higher efficiency.
% \fakeparagraph{}
We visualize networks found by \method in~\Cref{fig:mvcnn_breakdown}. The position of fusion module keeps changing given different latency and accuracy targets. Interestingly, the best accuracy is not strictly correlated with the largest model which further validates the necessity of searching the best model instead of increasing model size.

\section{Conclusion and Discussion}
\vspace{-1mm}
In summary, we introduce \method with SA-NAS for efficient partition and inference of ML models on-device with distributed smart sensors. The proposed SA-NAS approach enables end-to-end model search with flexible positioning of a splitting module for single-view and multi-view problems. The resource-constrained searching successfully identifies model architectures that reduce overall system latency, satisfy hardware constraints, while maintaining high accuracy. 
% The proposed method is evaluated with ImageNet for single-view and ModelNet40 for multi-view. 
Empirical results on ImageNet (single-view) and 3D Classification (multi-view) show that our approach can discover SOTA network architectures and model splitting solutions for distributed computing systems on AR/VR glasses. 

\textit{Limitations and future work.} The hardware model we adopt leverages an analytical model combing different hardware modalities, which are calibrated separately. Full system simulation with cycle accurate hardware models could be deployed for more precise latency evaluation.

\clearpage
{\small
\bibliographystyle{ieee_fullname}
\bibliography{egbib}

\begin{thebibliography}{10}
\bibitem{oculus}
Oculus quest 2: Our most advanced new all-in-one vr headset.
\end{thebibliography}
}

\clearpage
\appendix

% \begin{figure}[!t]
%     \centering
%     \includegraphics[width=\linewidth]{figs/modelnet-12views.pdf}
%     \caption{An example of 12 views of a 3D object in ModelNet40. For an object, 12 views are obtained by rotating the object every 30 degrees along the gravity direction.}
%     \label{fig:modelnet-12views}
% \end{figure}

% \section{Relation to Early Exiting Problem}
% \label{appendix:early_exiting}

\section{Multi-View Classification Dataset}
\label{appendix:mvcnn}
We also evaluate \method on a multi-view system with 3D Warehouse classification dataset that consists of 12,311 CAD models from 40 categories. CAD models in this benchmark come from 3D Warehouse~\cite{3dwarehouse}. For each object, 12 views are obtained by rotating the object every 30 degrees along the gravity direction.
% \Cref{fig:modelnet-12views} shows the 12 views sampled from an object. 
Using multiple views is necessary to obtain high classification accuracy according to prior work~\cite{feng2018gvcnn}.
We use the same training and testing split as in~\cite{wu20153d}. For our experiments, the reported metric is per-class accuracy.

% \usepackage{multirow}
\begin{table*}[h]
\resizebox{\linewidth}{!}{
\begin{tabular}{|cc|ccccc|}
\hline
\multicolumn{2}{|c|}{Input Resolution}                    & \multicolumn{5}{c|}{\{192, 224, 256, 288\}}                                                                                                                                                                                                                                                           \\ \hline
\multicolumn{1}{|c|}{Model Phase}              & Block    & \multicolumn{2}{c|}{NAS Search Space}                                           & \multicolumn{3}{c|}{SA-NAS Search Space}                                                                                                                                                                            \\ \hline
\multicolumn{1}{|c|}{Model Phase}              & Block    & \multicolumn{1}{c|}{Channel}               & \multicolumn{1}{c|}{Expansion Ratio} & \multicolumn{1}{c|}{\begin{tabular}[c]{@{}c@{}}Depth Before\\ Split Module\end{tabular}} & \multicolumn{1}{c|}{\begin{tabular}[c]{@{}c@{}}Depth After \\ Split Module\end{tabular}} &   \begin{tabular}[c]{@{}c@{}} Reduced\\ channel $d$ Module\end{tabular}                 \\ \hline
\multicolumn{1}{|c|}{-}                        & Conv     & \multicolumn{1}{c|}{\{16,24\}}           & \multicolumn{1}{c|}{\{1\}}           & \multicolumn{1}{c|}{-}                                                                   & \multicolumn{1}{c|}{-}                                                                   & -                             \\ \hline
\multicolumn{1}{|c|}{\multirow{2}{*}{Phase 1}} & MBConv-1 & \multicolumn{1}{c|}{\{16,24\}}           & \multicolumn{1}{c|}{\{1\}}           & \multicolumn{1}{c|}{\multirow{2}{*}{\{2,3,4,5\}}}                                        & \multicolumn{1}{c|}{\multirow{2}{*}{\{1,2,3\}}}                                          & \multirow{2}{*}{\{4,6,8\}}    \\ \cline{2-4}
\multicolumn{1}{|c|}{}                         & MBConv-2 & \multicolumn{1}{c|}{\{24,32\}}           & \multicolumn{1}{c|}{\{4,5,6\}}       & \multicolumn{1}{c|}{}                                                                    & \multicolumn{1}{c|}{}                                                                    &                               \\ \hline
\multicolumn{1}{|c|}{Phase 2}                  & MBConv-3 & \multicolumn{1}{c|}{\{32,40\}}           & \multicolumn{1}{c|}{\{4,5,6\}}       & \multicolumn{1}{c|}{\{1,2,3\}}                                                           & \multicolumn{1}{c|}{\{1,2,3\}}                                                           & \{6,8,10\}                    \\ \hline
\multicolumn{1}{|c|}{\multirow{2}{*}{Phase 3}} & MBConv-4 & \multicolumn{1}{c|}{\{64,72\}}           & \multicolumn{1}{c|}{\{4,5,6\}}       & \multicolumn{1}{c|}{\multirow{2}{*}{\{1,2,3\}}}                                          & \multicolumn{1}{c|}{\multirow{2}{*}{\{4,5,6,7,8,9\}}}                                    & \multirow{2}{*}{\{10,14,18\}} \\ \cline{2-4}
\multicolumn{1}{|c|}{}                         & MBConv-5 & \multicolumn{1}{c|}{\{112,120,128\}}     & \multicolumn{1}{c|}{\{4,5,6\}}       & \multicolumn{1}{c|}{}                                                                    & \multicolumn{1}{c|}{}                                                                    &                               \\ \hline
\multicolumn{1}{|c|}{\multirow{2}{*}{Phase 4}} & MBConv-6 & \multicolumn{1}{c|}{\{192,200,208,216\}} & \multicolumn{1}{c|}{\{6\}}           & \multicolumn{1}{c|}{\multirow{2}{*}{\{1,2,3,4\}}}                                        & \multicolumn{1}{c|}{\multirow{2}{*}{\{2,3,4,5,6\}}}                                      & \multirow{2}{*}{\{16,24,32\}} \\ \cline{2-4}
\multicolumn{1}{|c|}{}                         & MBConv-7 & \multicolumn{1}{c|}{\{216,224\}}         & \multicolumn{1}{c|}{\{6\}}           & \multicolumn{1}{c|}{}                                                                    & \multicolumn{1}{c|}{}                                                                    &                               \\ \hline
\multicolumn{1}{|c|}{-}                        & MBPool   & \multicolumn{1}{c|}{\{1792,1984\}}       & \multicolumn{1}{c|}{\{6\}}           & \multicolumn{1}{c|}{-}                                                                   & \multicolumn{1}{c|}{-}                                                                   & -                             \\ \hline
\end{tabular}
}
\caption{The supernet architecture configuration and search spaces for single-view \method on ImageNet.}
\label{appendix:tab:supernet}
\end{table*}

\section{Interlacing Concatenation for View Fusion}
\label{appendix:concat}

\begin{figure}[!t]
    \centering
    \includegraphics[width=\linewidth]{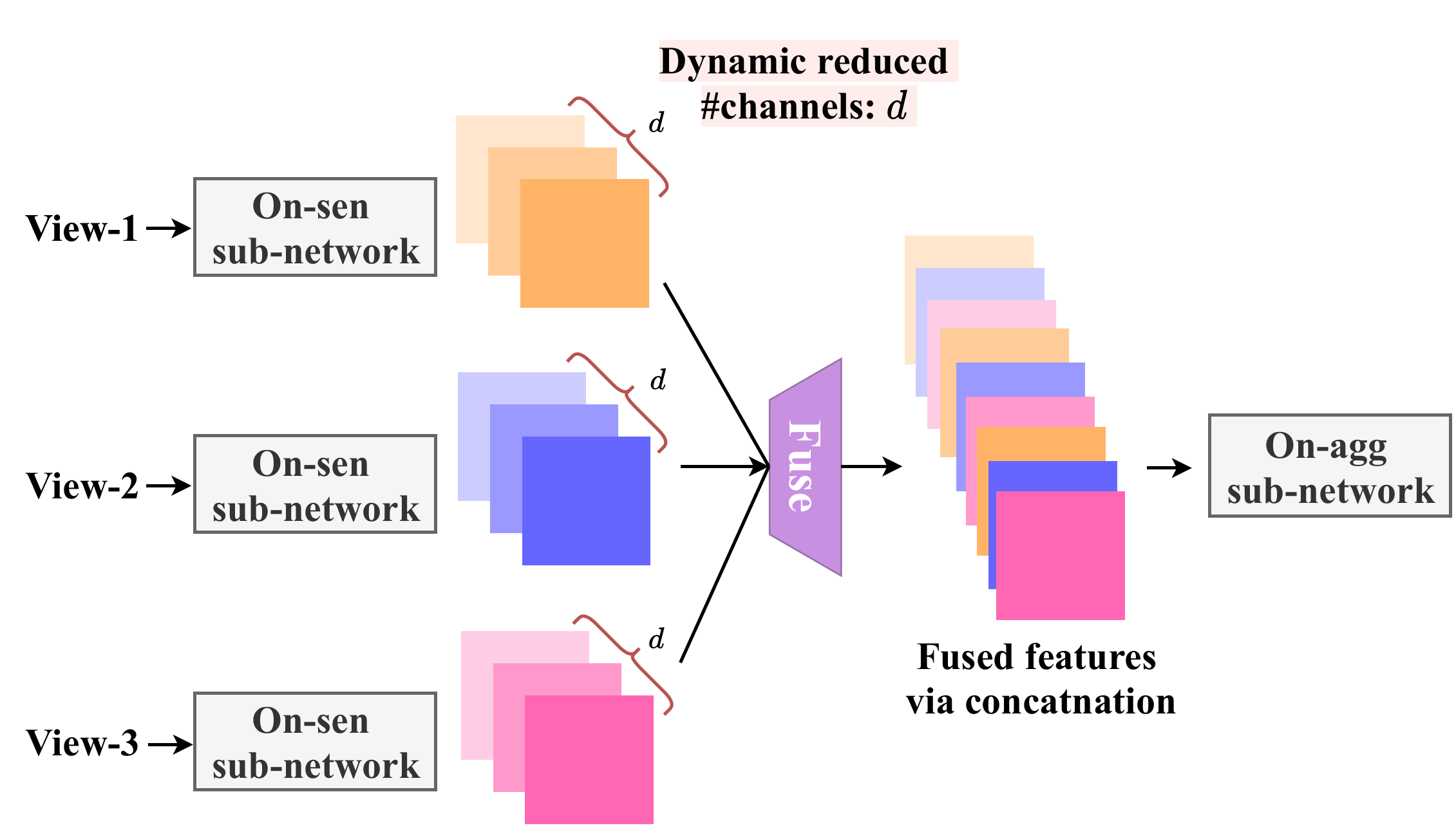}
    \caption{Illustration of interlacing concatenation. Interlacing concatenation guarantees that the $i$-th channel of the fused features is always from the $(i\ \textrm{mod}\ V)$-th view. In the example of this figure, $V=3$.}
    \label{fig:interlacing_concat}
\end{figure}

In~\Cref{sec:cm_arch}, we introduce a splitting module for multi-view systems, consisting of two layers \textit{Conv-Reduce} and \textit{View-Fuse}. Since we mainly focus on searching the position of \textit{View-Fuse}~(\ie, the position of splitting module), we introduce a simple fusion operation - concatenation. More specifically, we concatenate features from $V$ views in an interlacing way along the channel dimension as illustrated in~\Cref{fig:interlacing_concat}. In the interlacing concatenation, we first concatenate the first channel from $V$ views, then repeat the same operation for the second channels, and so on. We adopt interlacing concatenation because the number of channels of each view is dynamically sampled during training and interlacing concatenation guarantees that the $i$-th channel of the fused features is always from the $(i\ \textrm{mod}\ V)$-th view.

\section{Supernet Configuration}
\label{appendix:build_supernet}
As we mentioned in~\Cref{sec:split-aware-nas}, We build the supernet's search spaces mainly based on~\cite{dai2020fbnetv3,wang2021attentivenas} with some extra search spaces from our SA-NAS as shown in~\Cref{fig:SANAS}. Search spaces from standard NAS~\cite{dai2020fbnetv3,wang2021attentivenas} include number of layers, output channel size and expand ratio of each inverted residual block (MB). Following~\cite{dai2020fbnetv3,wang2021attentivenas}, we use Swish as activation function. The supernet architecture configuration and search space for single-view \method on ImageNet are summarized in~\Cref{appendix:tab:supernet}.

\section{Sampling Strategy of Splitting Modules for Supernet Training}
\label{appendix:training_supernet}

\subsection{Single-View \method}
\label{appendix:sample_singleview}
In supernet training, we jointly train multiple sub-networks sampled from the supernet at each iteration. Various sampling strategies can be used~\cite{cai2020once,yu2020bignas,dai2020fbnetv3,wang2021attentivenas}. For example, BigNAS~\cite{yu2020bignas} samples three kinds of networks: 1) the largest sub-network (``max''); 2) the smallest sub-network (``min''); 3) several randomly sub-sampled networks. Since the largest sub-network usually has better accuracy, it can be used as teacher to supervise other  sub-networks via knowledge distillation losses~\cite{hinton2015distilling}. Some variants of this ``sandwich'' sampling strategy are also proposed like sampling sub-networks according to their predicted accuracy~\cite{wang2021attentivenas}.

Compared with previous work, we have different interests on sub-networks. Eventually, we want to find the optimal architecture with exact one splitting module. However, as an information bottleneck, splitting module will inevitably introduce accuracy degradation. During training, we aim to minimize this degradation by improving the accuacy lower bound. We insert multiple splitting modules to help sub-networks increase tolerance of splitting modules. Inspired by BigNAS~\cite{yu2020bignas} and AttentiveNAS~\cite{wang2021attentivenas}, we sample five sub-networks per iteration to train jointly. 
\begin{enumerate}
    \item ``Max with zero hot'': The largest sub-network without splitting module can be treated as the best Pareto architecture. Training this sub-network helps improve the accuracy upper bound of all sub-networks. 
    \item ``Max with all hot'': The largest sub-network with all $N$ splitting modules contains all weights in the supernet. Training this sub-network ensures all weights are updated at least once per iteration. 
    \item ``Min with zero hot'': The smallest sub-network without splitting module contains weights which are shared across all sub-networks. Training this sub-network helps the optimization of the most frequently shared weights. 
    \item ``Min with all hot'': The smallest sub-network with all $N$ splitting modules can be treated as the worst Pareto architecture. Training this sub-network helps improve the accuracy lower bound of the all sub-networks.
    \item `Random with one hot'': A randomly sampled sub-network with exact one splitting module is the sub-network we are interested in during the second stage and final deployment.   
\end{enumerate}

\begin{figure}[!t]
    \centering
    \includegraphics[width=\linewidth]{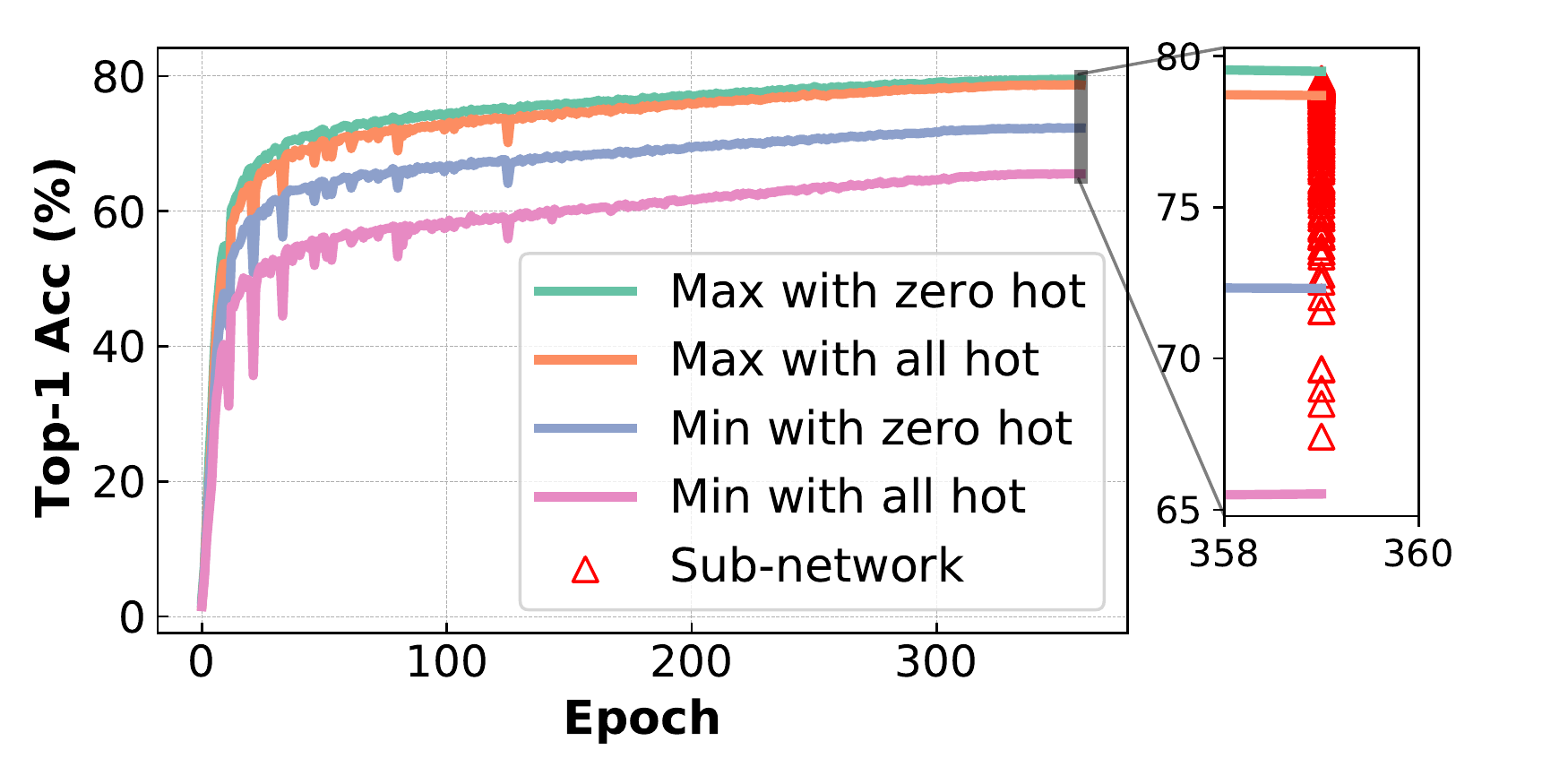}
    \caption{Training curves of four different sub-networks sampled from the supernet for single-view \method. ``max'' (or ``min'') indicts the maximum~(or ``minimum'') sub-network sampled from the supernet. ``all hot'' (or ``zero hot'') means that the sampled sub-network contains all possible (or none of) splitting modules. The drop of performance introduced (green line against orange line and  blue line against magenta line) by inserting splitting module is mitigated. The red triangles are sampled sub-networks with one splitting module (\ie, ``Random with one hot''). The green and pink lines can be treated as the upper and lower bounds of sub-networks.}
    \label{fig:imagenet_training}
\end{figure}

\begin{figure}[!t]
    \centering
    \includegraphics[width=\linewidth]{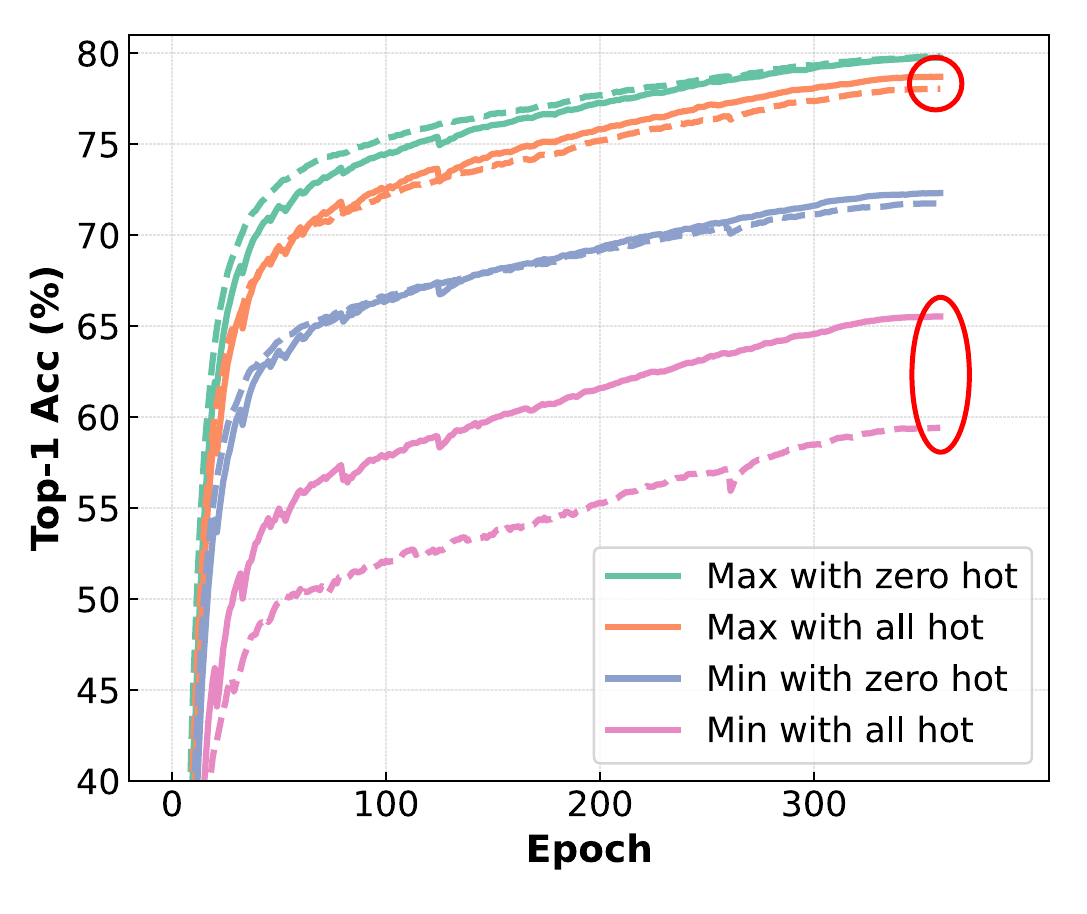}
    \caption{Comparison between our new sampling strategy~(solid lines) and baseline ``sandwich'' sampling strategy~(dashed lines) for single-view \method as  elaborated in~\Cref{appendix:sample_singleview}. Our new sampling strategy helps reduce accuracy drop in \method due to splitting modules. For example, the discrepancy between solid blue and pink lines is significantly smaller than discrepancy between dashed blue and pink lines.}
    \label{fig:subnet_comp}
\end{figure}

We visualize training curves of aforementioned sub-networks in~\Cref{fig:imagenet_training}. The red triangles are some sampled sub-networks with one splitting module. Their accuracy is bounded by the ``Max with zero hot'' and ``Min with all hot'' sub-networks. 

In addition, we compare our new sampling strategy against the baseline ``sandwich'' sampling strategy, which replaces ``Max with all hot'' (and ``Min with all hot'') with ``Max with one hot'' (and ``Min with one hot'') during supernet training. From~\Cref{fig:subnet_comp}, we find that sampling more than one splitting modules during training helps reduce accuracy drop introduced by splitting modules in \method. For example, the discrepancy between solid blue and pink lines is significantly smaller than discrepancy between dashed blue and pink lines.

\subsection{Multi-View \method}

For multi-view \method, each splitting module contains a fusion operation which can only be performed once for one network. So we are not able to sample more than one splitting modules during neither supernet training nor architecture searching. We use the baseline ``sandwich'' sampling strategy for multi-view \method. 

\section{Analysis of Different Initializations}
\label{appendix:kaiming_init}
As we discussed in~\Cref{sec:cm_arch}, both forward and backward passes of a convolution layer can be expressed by convolution operations. The goal of initialization is to ensure the magnitude of output~(and gradient) does not explode during forward~(and backward) pass as shown in the following two equations,
\begin{align}
&\texttt{CONV}_\mathrm{forward}\left(\rmW, \rvx_{l}\right)=\rvy_l \sim \mathcal{N}(0,1)
\label{equ:kaiming_forward}\\
&\texttt{CONV}_\mathrm{backward}\left(\rmW^T, \frac{\partial \mathcal{L}}{\partial \rvy_l}\right)=\frac{\partial \mathcal{L}}{\partial \rvx_{l}}\sim \mathcal{N}(0,1),
\label{equ:kaiming_backward}
\end{align}
where we assume the activation function as ReLU~\cite{he2016deep}, $\rvx_{l+1} = \operatorname{max}(\rvy_l, 0)$.
Solving~\Cref{equ:kaiming_forward} or~\Cref{equ:kaiming_backward} leads to Kaiming Fan-In or Kaiming Fan-Out~\cite{he2015delving},
\begin{align}
    \rmW &\sim \mathcal{N}\left(0, \frac{2}{k^2\cdot c_\mathrm{in}}\right)\quad\quad \textsc{(Fan-In)} \\
    \rmW &\sim \mathcal{N}\left(0, \frac{2}{k^2\cdot c_\mathrm{out}}\right)\quad\ \  \textsc{(Fan-Out)},
\end{align}
where $k$ is kernel size and $c_\mathrm{in}$~(and $c_\mathrm{out}$) is the input~(and output) channel size. 

In a splitting module, the difference between input and output channel sizes is usually enormous. For example, in a certain \textit{Conv-Reduce}, we have $c_\mathrm{in}=256$ and $c_\mathrm{out}=8$. In this case, Fan-In mode's weights variance is $\frac{256}{8}=32$ times larger than Fan-Out's. Choosing either Fan-In or Fan-Out will cause the other one's variance too large or too small. Although Xavier~\cite{glorot2010understanding}'s arithmetic average of $c_\mathrm{in}$ and $c_\mathrm{out}$ may mitigate this issue, it is far from enough because arithmetic average between two numbers is dominated by the larger one, $0.5\cdot(256+8)\gg 8$. 

Our split-aware initialization adopts geometric average instead of arithmetic average to make a better balance between forward and backward, $\sqrt{c_\mathrm{in}\cdot c_\mathrm{out}}$. In the next section, we empirically show that supernet training can benefit from our split-aware initialization.

\section{Empirical Comparison of Different Initializations}
\label{appendix:init_numbers}

% \begin{table}[h]
%     \centering
%     \begin{tabular}{c|cc}
%     \toprule
%         Representative Networks & Kaiming~\cite{he2015delving} &  Ours \\
%     \midrule
%         Min with all hot & 65.66 \\ 
%         Min with zero hot & 72.20 \\
%         Max with all hot & 78.676 \\
%         Max with zero hot & 79.462 \\ 
%     \bottomrule
%     \end{tabular}
%     \caption{Top-1 accuracy comparison of Kaiming uniform~(Fan-In) and our split-aware initialization~(\Cref{sec:cm_arch})}
%     \label{tab:init_acc_compare}
% \end{table}

\begin{figure}
    \centering
    \includegraphics[width=0.9\linewidth]{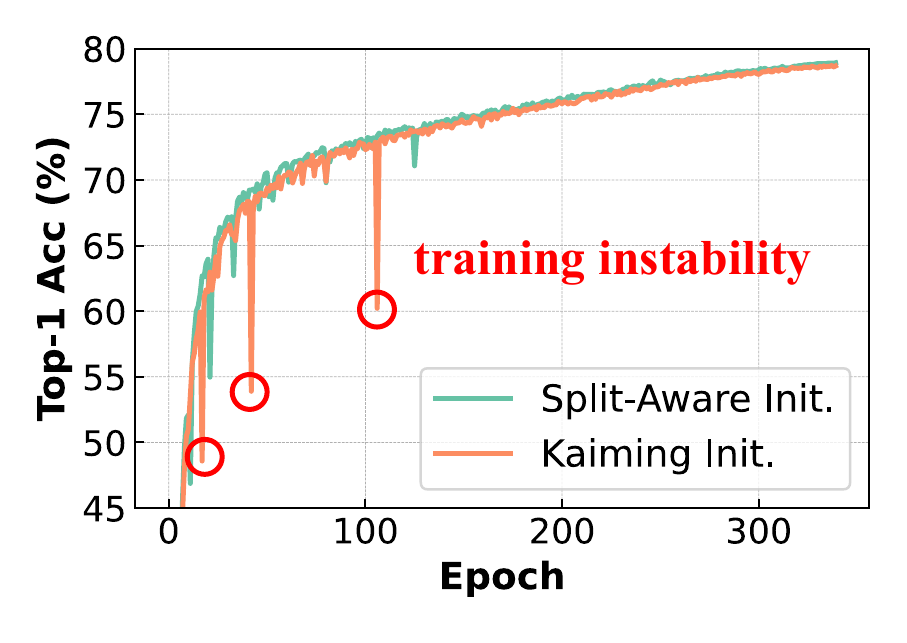}
    \caption{Accuracy comparison between Kaiming and our split-aware initalizations on the ``Max with all hot'' sub-network.}
    \label{fig:init_comp_result}
\end{figure}

We compare our split-aware initialization against Kaiming initialization (Fan-In) on ``Max with all hot'' sub-network's accuracy for ImageNet in~\Cref{fig:init_comp_result}. Our split-aware initialization improves training stability as well as final accuracy (by~0.3\%). In addition, if the base learning rate~(0.05 in~\Cref{fig:init_comp_result}) is slightly increased, conventional initialization will lead to divergence in training. 

\section{Hardware Modeling}
\label{appendix:hardware_modeling}
As discussed in~\Cref{sec:stage-two}, we use a hardware simulator customized for a realistic head mounted device. The on-sen. processor is equipped with a 16nm neural processing unit (NPU) with peak performance of $\operatorname{Comp}_\texttt{sen}=125$ GOP/s. So, the latency of on-sensor part (in~\Cref{equ:objective}) can thus be approximately computed via $T_\texttt{sen}(f_\texttt{sen}, \rvx) = \operatorname{OP}(f_\texttt{sen}, \rvx)/\operatorname{Comp}_\texttt{sen}$, where $\operatorname{OP}(\cdot,\cdot)$ is the profiling function through cycle-accurate simulation for measuring the number of operations given model and input. In addition, the on-sen. processor's peak memory is $\operatorname{Mem}_\texttt{sen}=2$ MB. Thus, the peak memory consumption of the on-sen part cannot exceed this peak memory constraint $M(f_\texttt{sen}, \rvx)\leq\operatorname{Mem}_\texttt{sen}$. When computing the peak memory consumption ($M(\cdot, \cdot)$), we consider the memory of both weights and activations: $M(f_\texttt{sen}, \rvx)=M_\texttt{W}(f_\texttt{sen})+M_\texttt{A}(f_\texttt{sen}, \rvx)$, where $M_\texttt{W}(f_\texttt{sen})$ is the memory consumption for storing all weights of $f_\texttt{sen}$ and $M_\texttt{A}(f_\texttt{sen}, \rvx)$ measures the peak memory consumption of activation taking residual connections into consideration. In this work, we consider homogeneous sensors which can represent most of {AR/VR} devices like {Quest2}~\cite{oculus}. We leave the extension of \method as a future work when heterogeneous sensors occur.

\section{Adaptability of \method}
\label{appendix:imagenet_network_change}

\begin{figure}
    \centering
    \includegraphics[width=\linewidth]{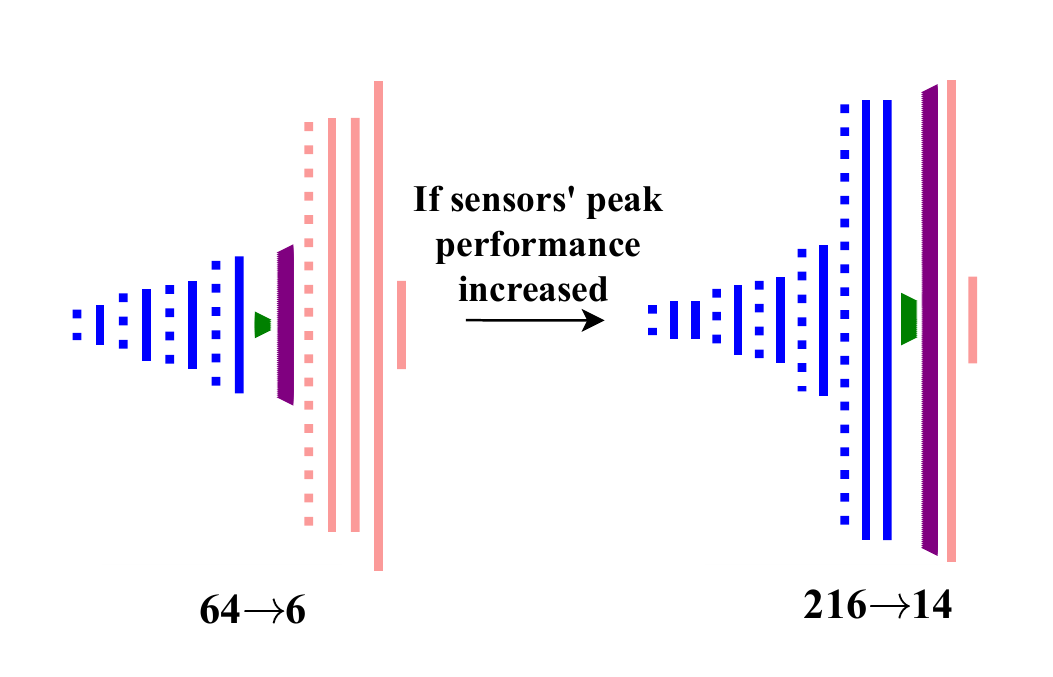}
    \caption{Evolution of the best architectures when on-sen. processor's peak computation performance increases. SA-NAS can automatically put more and wider layers on sensors. }
    \label{fig:hardware_change_mvcnn}
\end{figure}

As we discussed in~\Cref{sec:split-aware-nas}, two-stage NAS decouples the supernet training  (stage 1) and architecture searching (stage 2). As a result, when the hardware configuration changes, one just needs to rerun the stage 2 without retraining the supernet. In this section, we change the hardware configuration and observe how the searched architectures evolve to fit the change of hardware configuration. Specifically, we increase the computation capability of on-sen. processors by four times and show the change of architectures with the best accuracy for the multi-view task in~\Cref{fig:hardware_change_mvcnn}. The left architecture is the best network for the default hardware configuration ($\operatorname{Comp}_\texttt{sen}=125$ GOP/s) from~\Cref{tab:more_models_transfer}~(\method-C). 
If on-sen. processors' peak computation performance is increased by 4$\times$, SA-NAS can automatically put more and wider layers on-sen. and reduce the number of on-agg. layers to make a better trade-off. The network on the right achieves a better performance~(94.0\% top-1 accuracy) and lower latency~(0.62 ms) compared with the left one.

% \section{Multi-View Classification}
% \label{appendix:mvcnn}

\end{document}